\documentclass{article}

\usepackage{colm2024_conference}
\usepackage[edges]{forest}
\usepackage[utf8]{inputenc} 
\usepackage[T1]{fontenc}    
\usepackage{hyperref}       
\usepackage{url}            
\usepackage{booktabs}       
\usepackage{amsfonts}       
\usepackage{nicefrac}       
\usepackage{microtype}      
\usepackage{xcolor}         
\usepackage{amsmath} 
\usepackage{graphicx}
\usepackage{multirow}
\usepackage{comment}
\usepackage{tikz}
\usepackage{pgfplots}
\usepackage{pgfplotstable}
\usepackage{makecell}
\usepackage{adjustbox}
\usepackage{pifont}
\usepackage{float}
\usepackage{graphicx}
\usepackage{pdflscape}
\usepackage{setspace}
\usepackage{booktabs}
\usepackage{colortbl}

\definecolor{Gray}{gray}{0.94}
\definecolor{DarkGreen}{RGB}{30,130,30}

\newcommand{\xmark}{\textcolor{red}{\ding{55}}}
\newcommand{\cmark}{\textcolor{DarkGreen}{\ding{51}}}

\title{Scaling of Search and Learning: A Roadmap to Reproduce o1 from Reinforcement Learning Perspective}

\colmfinalcopy

\author{
Zhiyuan Zeng$^{1}$\thanks{Equal contribution. Listing order is random.} \hspace{.1em}
Qinyuan Cheng$^{1*}$ \hspace{.1em}
Zhangyue Yin$^{1*}$ \hspace{.1em}
Bo Wang$^{1*}$ \hspace{.1em}
\\
\textbf{
Shimin Li$^{1}$ \hspace{.1em}
Yunhua Zhou$^{2}$ \hspace{.1em}
Qipeng Guo$^{2}$ \hspace{.1em}
Xuanjing Huang$^{1}$ \hspace{.1em}
Xipeng Qiu$^{1}$\thanks{Corresponding author.
 Correspondence to: \texttt{\{cengzy23,yinzy21,bwang22\}@m.fudan.edu.cn} \texttt{chengqy2019@foxmail.com} \texttt{\{xjhuang, xpqiu\}@fudan.edu.cn}}
}
\\
[1ex]
$^{1}$Fudan University \\
$^{2}$Shanghai AI Laboratory \\
}

\fancypagestyle{firstpage}{
  \lhead{\begin{picture}(0,0)\put(0,-6){\includegraphics[width=0.3\linewidth]{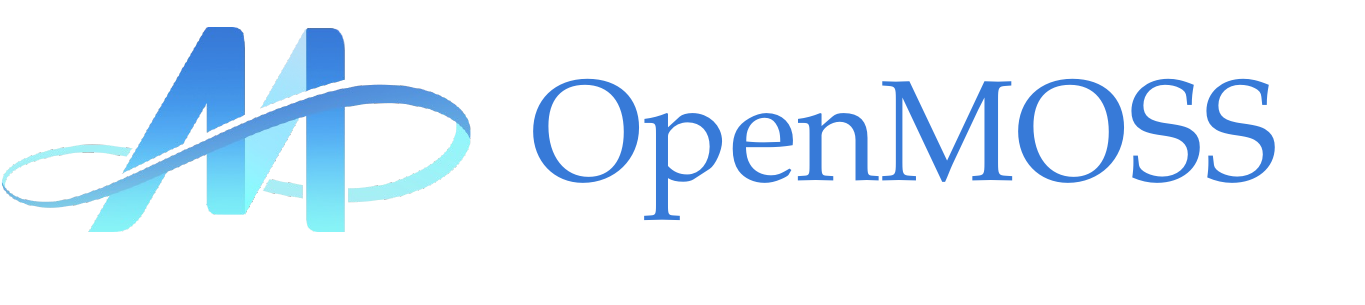}}\end{picture}}
}

\newcommand{\blackxmark}{\textcolor{black}{\ding{55}}}

\begin{document}

\maketitle
\thispagestyle{firstpage}

\begin{abstract}

OpenAI o1 represents a significant milestone in Artificial Inteiligence, which achieves expert-level performances on many challanging tasks that require strong reasoning ability.
OpenAI has claimed that the main techinique behinds o1 is the reinforcement learining \citep{o1_blog, o1_system_card}.
Recent works use alternative approaches like knowledge distillation to imitate o1's reasoning style, but their effectiveness is limited by the capability ceiling of the teacher model.
Therefore, this paper analyzes the roadmap to achieving o1 from the perspective of reinforcement learning, focusing on four key components: policy initialization, reward design, search, and learning.
Policy initialization enables models to develop human-like reasoning behaviors, equipping them with the ability to effectively explore solution spaces for complex problems.
Reward design provides dense and effective signals via reward shaping or reward modeling, which is the guidance for both search and learning.
Search plays a crucial role in generating high-quality solutions during both training and testing phases, which can produce better solutions with more computation. 
Learning utilizes the data generated by search for improving policy, which can achieve the better performance with more parameters and more searched data.
Existing open-source projects that attempt to reproduce o1 can be seem as a part or a variant of our roadmap.
Collectively, these components underscore how learning and search drive o1's advancement, making meaningful contributions to the development of LLM.
\end{abstract}

\begin{figure}[h]
    \centering
    \includegraphics[width=0.7\linewidth]{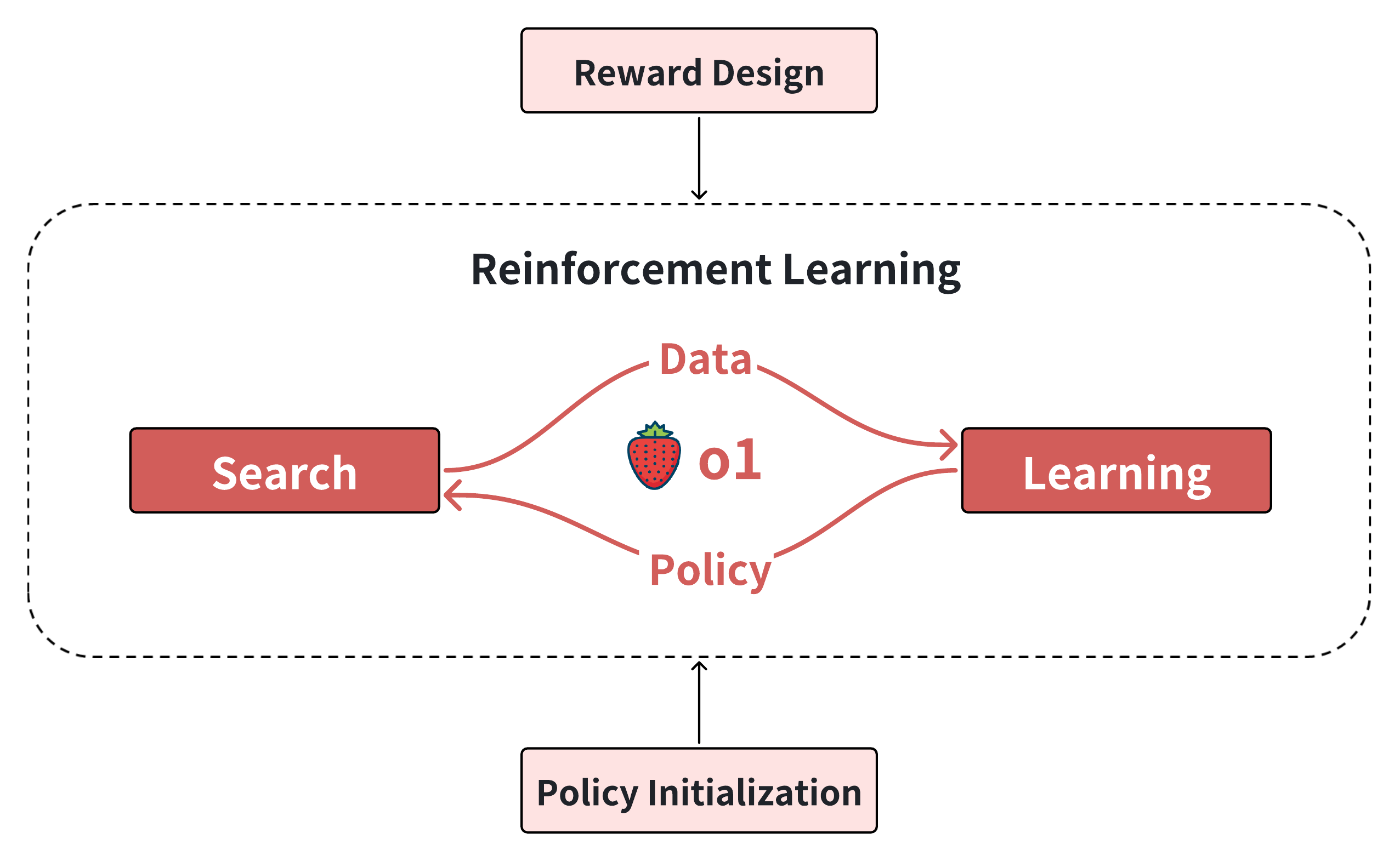}
    \vspace{-1em}
    \caption{The overview of this roadmap including policy initialization, reward design, search and learning.}
    \label{fig:roadmap}
\end{figure}

\newpage

\definecolor{hidden-draw}{RGB}{0,0,0}

\definecolor{coral}{RGB}{255, 127, 80}      
\definecolor{sage}{RGB}{128, 177, 133}      
\definecolor{azure}{RGB}{91, 155, 213}      
\definecolor{lilac}{RGB}{180, 160, 207}     

\tikzstyle{leaf}=[draw=hiddendraw,
    rounded corners,minimum height=1.2em,
    fill=hidden-orange!40,text opacity=1,    align=center,
    fill opacity=.5,  text=black,align=left,font=\scriptsize,
inner xsep=3pt,
inner ysep=1pt,
]

\begin{figure}[thp]
\centering
\begin{forest}
    forked edges,
    for tree={
        grow=east,
        reversed=true,
        anchor=base west,
        parent anchor=east,
        child anchor=west,
        base=left,
        font=\footnotesize,
        rectangle,
        draw,
        rounded corners,align=left,
        minimum width=2.5em,
        minimum height=1.2em,
        edge+={darkgray, line width=0.6pt},
        s sep=6pt,
        l sep=10pt,
        inner xsep=3pt,
        inner ysep=1pt,
        ver/.style={rotate=90, child anchor=north, parent anchor=south, anchor=center},
    },
    where level=1{text width=3.6em}{},
    where level=2{font=\scriptsize}{},
    where level=3{font=\scriptsize}{},
    where level=4{font=\scriptsize}{},
    where level=5{font=\scriptsize}{},
    [Roadmap to o1, draw=gray, color=gray!100, fill=gray!15, thick, text=black, ver
        [Policy Init, color=coral, fill=coral!15, thick, text=black
            [Pre-Training, color=coral, fill=coral!15
                [Language Understanding\\ and Generation, color=coral, fill=coral!15
                    [
                        \citet{Radford2018ImprovingLU,manning2022human}
                    ]
                ]
                [World Knowledge, color=coral, fill=coral!15
                    [
                        \citet{Radford2019LanguageMA,brown2020languagemodelsfewshotlearners}
                    ]
                ]
                [Basic Reasoning, color=coral, fill=coral!15
                    [
                        \citet{lewkowycz2022solving,Thought_Propagation}
                    ]
                ]
            ]
            [Instruction Fine-Tuning, color=coral, fill=coral!15
                [
                    \citet{Survey_Instruction_Tuning,Instruction_Pre_Training}
                ]
            ]
            [Human-like Reasoning\\Behaviors, color=coral, fill=coral!15
                [Goal Clarification, color=coral, fill=coral!15
                    [
                        \citet{kondrakunta2018toward,Proactive_Dialogues}
                    ]
                ]
                [Task Decomposition, color=coral, fill=coral!15
                    [
                        \citet{Least-to-Most,Complex_Tasks_Compositional}
                    ]
                ]
                [Alternative Proposal, color=coral, fill=coral!15
                    [
                        \citet{jintelligence9020023,Divergent_Thinking}
                    ]
                ]
                [Solution Generation, color=coral, fill=coral!15
                    [
                        \citet{wei2022chain,Zero_shot_CoT}
                    ]
                ]
                [Self-Evaluation, color=coral, fill=coral!15
                    [
                        \citet{AnthropicCAI,Self-Verification}
                    ]
                ]
                [Self-Correction, color=coral, fill=coral!15
                    [
                        \citet{Intrinsic_Self_Correction,Learning_to_Check}
                    ]
                ]
                ]
            ]
        [Reward, color=sage, fill=sage!15, thick, text=black
            [Granularity, color=sage, fill=sage!15
                [Outcome, color=sage, fill=sage!15
                    [
                        \citet{OpenAIMathVerifierORM, shao2024deepseekmathpushinglimitsmathematical}
                    ]
                ]
                [Process, color=sage, fill=sage!15
                    [
                        \citet{VerifySbyS,MathShepherd}
                    ]
                ]
            ]
            [Methods, color=sage, fill=sage!15
                [Realistic, color=sage, fill=sage!15
                    [
                        \citet{StepCoder,MineDojo,Alfworld}
                    ]
                ]
                [Learned, color=sage, fill=sage!15
                    [
                        \citet{DeepRLFromHF, FromRtoQ}
                    ]
                ]
            ]
            [Generalization, color=sage, fill=sage!15
                [Reward Ensemble, color=sage, fill=sage!15
                    [
                        \citet{AlphaGo, VerifySbyS}
                    ]
                ]
                [World Model, color=sage, fill=sage!15
                    [
                        \citet{WorldModel,NavigationWorldModel}
                    ]
                ]
            ]
        ]
        [Search, color=azure, fill=azure!15, thick, text=black
            [Guidance, color=azure, fill=azure!15
                [Internal, color=azure, fill=azure!15
                    [Model Uncertainty, color=azure, fill=azure!15
                        [
                            \citet{SC,SemanticUncertainty}
                        ]
                    ]
                    [Self-evaluation, color=azure, fill=azure!15
                        [
                            \citet{Self_refine, Self_rewarding}
                        ]
                    ]
                ]
                [External, color=azure, fill=azure!15
                    [Environmental Feedback, color=azure, fill=azure!15
                        [
                            \citet{OpenAIMathVerifierORM,VerifySbyS}
                        ]
                    ]
                    [Heuristic Rules, color=azure, fill=azure!15
                        [
                            \citet{yao2023tree,RAP}
                        ]
                    ]
                ]
                [External+Internal, color=azure, fill=azure!15
                    [Value Function, color=azure, fill=azure!15
                        [
                            \citet{OVM, AlphaMath}
                        ]
                    ]
                    [Uncertainty+Verifier, color=azure, fill=azure!15
                        [
                            \citet{MathShepherd, Scaling_test_time_compute}
                        ]
                    ]
                ]
            ]
            [Strategy, color=azure, fill=azure!15
                [Tree Search, color=azure, fill=azure!15
                    [Best-of-N, color=azure, fill=azure!15
                        [
                            \citet{Speculative_bon, Variational_bon}
                        ]
                    ]
                    [Beam Search, color=azure, fill=azure!15
                        [
                            \cite{Self_evaluation_guided_beam_search, OVM}
                        ]
                    ]
                    [MCTS, color=azure, fill=azure!15
                        [
                            \citet{Alpha-zero-like, ppo_mcts}
                        ]
                    ]
                ]
                [Sequential Revisions, color=azure, fill=azure!15
                    [
                        \citet{Self_refine, SCoRE}
                    ]
                ]
            ]
        ]
        [Learning, color=lilac, fill=lilac!15, thick, text=black
            [Policy Gradient, color=lilac, fill=lilac!15
                [REINFORCE, color=lilac, fill=lilac!15
                    [
                        \citet{reinforce,remax}
                    ]
                ]
                [PPO, color=lilac, fill=lilac!15
                    [
                        \citet{shao2024deepseekmathpushinglimitsmathematical,rlhf-secret-1}
                    ]
                ]
                [DPO, color=lilac, fill=lilac!15
                    [
                        \citet{mcts-dpo,svpo}
                    ]
                ]
            ]
            [Behavior Cloning, color=lilac, fill=lilac!15
                [
                    \citet{Star,AlphaMath}
                ]
            ]
        ]
    ]
\end{forest}
\caption{Implementation of the policy initialization, reward design, search, and reinforcement learning.}
\label{fig:o1-roadmap}
\end{figure}
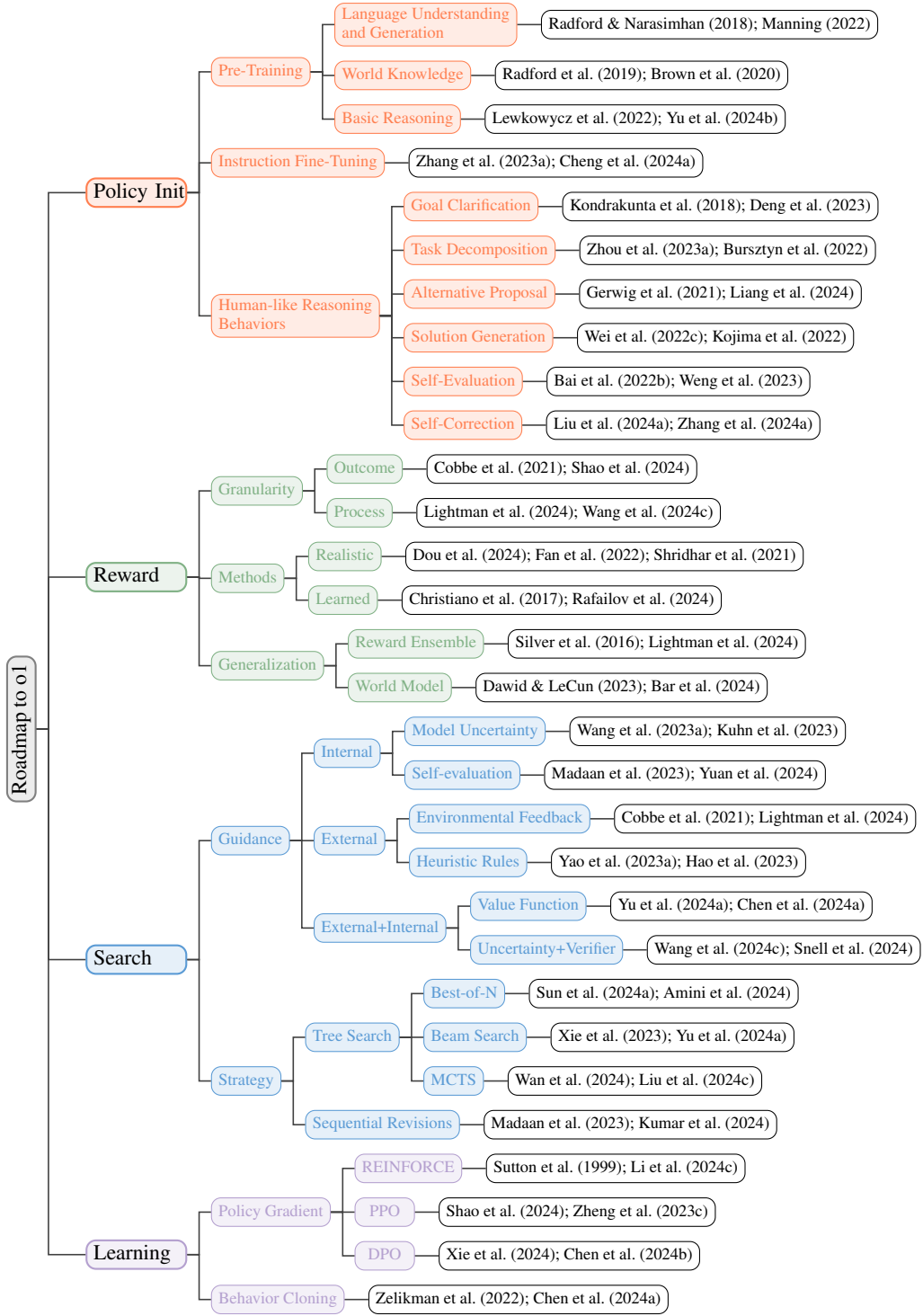

\newpage

\begin{quote}
\textit{One thing that should be learned from the bitter lesson is the great power of general purpose methods, of methods that continue to scale with increased computation even as the available computation becomes very great. The two methods that seem to scale arbitrarily in this way are \textbf{search} and \textbf{learning}.}
\hfill --- \textit{Richard Sutton, 2019}
\end{quote}

\section{Introduction}
\label{sec:introduction}

The field of Artificial Intelligence (AI) has witnessed unprecedented exploration and advancement of Large Language Models (LLMs) over the past two years. LLMs have progressively evolved to handle increasingly sophisticated tasks such as programming and solving advanced mathematical problems. OpenAI o1 represents a significant milestone in AI, which can generate very long reasoning process and conduct human-like reasoning actions like clarifying and decomposing questions, reflecting and correcting previous mistakes, exploring new solutions when encountering failure modes.
The o1 model dramatically transcended the reasoning capabilities of preceding LLMs, achieving performance comparable to PhD-level proficiency.
Its remarkable reasoning achievements signify OpenAI's progression to the second stage (``Reasoner'') in their five-stage roadmap to Artificial General Intelligence (AGI).

The blog and system card of o1 demonstrate that the performance of o1 consistently improves with increasing the computation of reinforcement learning and inference~\citep{o1_blog,o1_system_card}. This suggests that o1 could drive two paradigm shifts in AI: from (self-)supervised learning toward reinforcement learning, and from scaling solely training computation to scaling both training and inference computation.

o1 scales up the train-time compute with reinforcement learning and the test-time compute with more thinking. We take search as a way to implement the thinking process of o1, since search is scalable~\citep{the_bitter_lesson} and there are many successful researches that use search for training and decision in reinforcement learning, like AlphaGo~\citep{AlphaGo} and AlphaGo Zero \citep{AlphaZero}. In this paper, we take reinforcement learning as the core in the roadmap to o1. Our roadmap is illustrated in Figure~\ref{fig:roadmap} and consists of four components: policy initialization, reward design, search, and learning.
We believe these four parts are the keys to constructing LLM with strong reasoning abilities like o1.

As depicted in Figure~\ref{fig:o1-roadmap}, our roadmap starts with policy initialization. In the context of LLMs, the policy ($\pi(a|s)$) typically refers to the probability distribution for generating the next token/step/response (action) based on a given context (state). Policy initialization brings human-like reasoning behaviors to LLMs, like task composition, self-evaluation and self-correction.
Next, we enter the reward design, which aims to provide guiding signals for search and learning.
The reward design can take or reshape the reward signal from the environment or learn a reward model from preference data. Both policy initialization and reward designs are preparations for search and learning. Search plays an important role in generating high-quality solutions at both training and testing phases, which can produce better solutions with more computation. Learning utilizes the data generated by search for improving the policy. The data used for learning is derived from the interaction of the LLM with the environment, rather than being manually curated by human experts, therefore eliminating the need for costly data annotation and enabling the potential for achieving superhuman performance.

\paragraph{Policy Initialization}
Training an LLM from scratch using reinforcement learning is exceptionally challenging due to its vast action space. Fortunately, we can leverage extensive internet data to pre-train a language model, establishing a potent initial policy model capable of generating fluent language outputs.
Moreover, prompt engineering and supervised fine-tuning help models acquire human-like reasoning behaviors, enabling them to think systematically and validate their own results. These approaches enable models to thoroughly explore their solution spaces, leading to more comprehensive problem-solving capabilities.

\paragraph{Reward Design} Both search and learning require guidance from reward signals to improve the policy. There are different levels of action granularity, each corresponding to varing levels of reward signals granularity, which can be explored further. Additionally, these signals are often sparse or even nonexistent in many environments. To transform sparse outcome reward to dense process reward, there are some reward shaping methods \citep{PotentialFunc}. For the environment where the reward signal is unavailable, like the task of story writing, we can learn a reward model from preference data \citep{rlhf-anthropic} or expert data \citep{IRL}.The construction of reward model can further evolve into building a world model \citep{WorldModel}.

\paragraph{Search} Search plays a crucial role during both the training and testing phases. The training time search refers to generating training data from search process. The advantage of using search to generate training data, as opposed to simple sampling, is that search yields better actions or solutions—i.e., higher-quality training data—thereby enhancing learning effectiveness.
During inference, search continues to play a vital role in improving the model's sub-optimal policies. For instance, AlphaGo \citep{Alpha-zero-like} employs Monte Carlo Tree Search (MCTS) during testing to enhance its performance. However, scaling test-time search may lead to inverse scaling due to distribution shift: the policy, reward, and value models are trained on one distribution but evaluated on a different one~\citep{RewardOverOptim}.

\paragraph{Learning} Learning from human-expert data requires costly data annotation. In contrast, reinforcement learning learns through interactions with the environment, eliminating the need for expensive data annotation and offering the potential for superhuman performance. In this roadmap, reinforcement learning utilizes data generated by search for learning via policy gradient or behavior cloning. Policy gradient methods have high data utilization, as they leverage both positive and negative solutions, whereas behavior cloning is advantageous in terms of simplicity and memory efficiency. A prominent example of the iterative interaction between search and learning is AlphaGo Zero ~\citep{AlphaZero}, which combines Monte Carlo Tree Search (MCTS) ~\citep{metropolis1949monte} as the search algorithm with behavior cloning as the learning method, ultimately achieving superhuman performance in the game of Go.

We provide a detailed exploration of the potential implementations of Policy Initialization (Section \ref{policy_initilization}), Reward Design (Section \ref{sec:reward_modeling}), Search (Section \ref{search}), and Learning (Section \ref{learning}). Additionally, we review existing open-source o1 projects, illustrating how they may either serve as components of our framework or as specific instances within it (Section \ref{open-source-o1}). Finally, we discuss the future development trends of o1 and the associated challenges (Section \ref{future-directions}).

\section{Background}
\label{sec:background}
\begin{figure}
    \centering
    \includegraphics[width=1.0\linewidth]{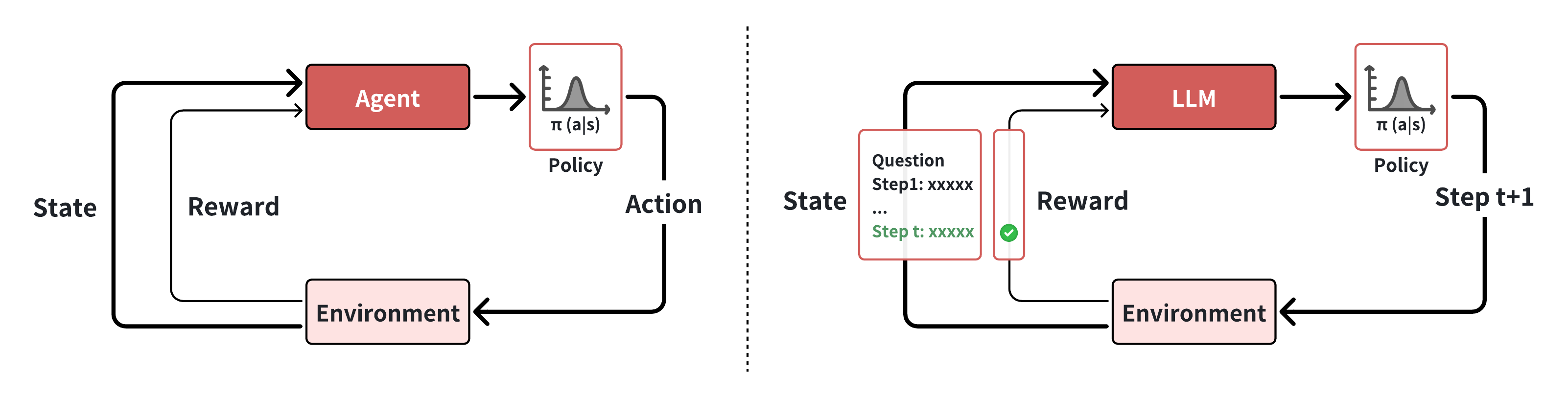}
    \vspace{-1em}
    \caption{The visualization of the interaction between agent and einvironment in reinforcement learning for LLMs. Left: traditional reinforcement learning. Right: reinforcement learning for LLMs. The figure only visualizes the step-level action for simplicity. In fact, the action of LLM can be either token-, step-, or solution-level.}
    \label{fig:rl_back}
    \vspace{-1em}
\end{figure}

Since this roadmap is designed from perspective of reinforcement learning, we introduce some background of reinforcement learning and its connection to LLM in this section. Unlike other learning paradigms, reinforcement learning learns through interaction with the environment, rather than learning from a static training dataset. In reinforcement learning, an agent learns by receiving rewards from the environment as it explores. Figure \ref{fig:rl_back} illustrates the interaction between agent and environment in reinforcement learning for LLM.

\paragraph{Agent}

Agent is the entity that interacts with environment, which makes decision according to its policy. Formally, a \textbf{policy} \( \pi \) is a mapping from states to actions. It is often represented as a probability distribution (\( \pi(a|s) \)) over actions given a state \( s \), where the agent selects actions based on these probabilities.  

In the context of LLMs, an \textbf{agent} refers to LLM itself, its policy specify the probability distribution of either token-, step-, or solution-level actions based on the current state. The \textbf{state} \( s_t \) consists of the input provided to the model at time \( t \), including both user inputs and the model’s earlier outputs. The \textbf{action} taken by the model can vary depending on the problem setting; it involves generating a single token, completing a step, or providing a solution.

\paragraph{Environment}
Environment refers to the system or world outside the agent. It responds to agent's actions and provide feedback in terms of next state $s_{t+1}$ and rewards $r(s_t,a_t)$.

Environmental feedback can be categorized as either deterministic or stochastic. Stochastic feedback is characterized by a transition distribution \( p(s_{t+1}, r_{t+1} | s_t, a_t) \), as seen in systems like dialogue models, where user responses are inherently unpredictable. On the other hand, deterministic feedback involves no randomness, yielding a fixed next state \( s_{t+1} \) and reward \( r(s_t, a_t) \). For instance, when a LLM solves a mathematical problem, the transition is deterministic, where the current state \( s_t \) and action \( a_t \) are combined to produce the next state \( s_{t+1} \).
 
\section{Policy Initialization}
\label{policy_initilization}

In reinforcement learning, a policy defines how an agent selects actions in response to environmental states. As discussed in Section~\ref{sec:background}, LLMs operate with actions at three granularity levels: solution-level, step-level, and token-level. Solution-level actions represent the coarsest granularity, treating the entire solution as a single action. Step-level actions operate at an intermediate granularity, where individual steps serve as discrete actions. Token-level actions provide the finest granularity, treating each individual token as an action. Taking token-level actions as an example, the action space contains thousands of tokens from the vocabulary, establishing a well-initialized policy becomes essential for effective model performance~\citep{brown2020languagemodelsfewshotlearners}.

\begin{figure}[t]
    \centering
    \includegraphics[width=\linewidth]{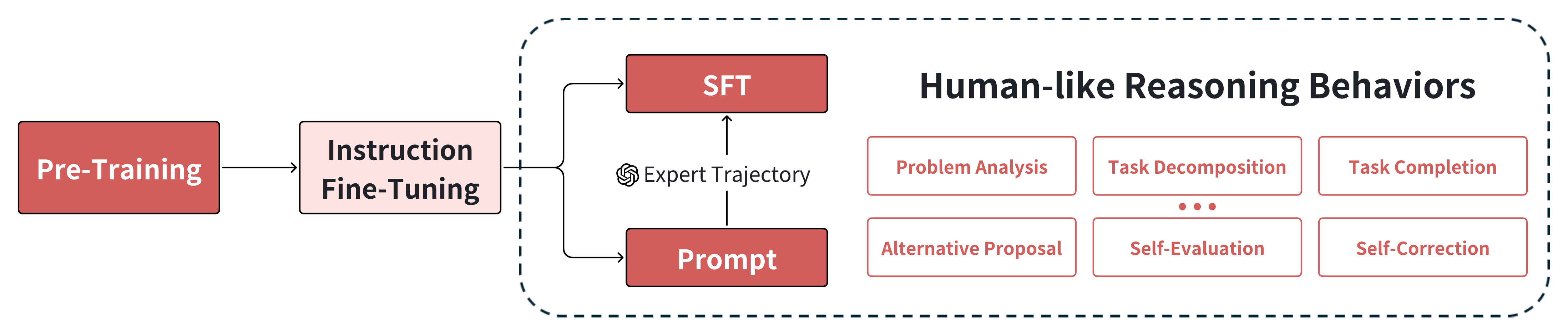}
    \vspace{-1em}
    \caption{The process of policy initialization, including pre-training, instruction fine-tuning, and the injection of human-like reasoning behaviors. These behaviors can be learned through SFT or triggered by prompts.
}
\label{fig:policy_initialization}
\end{figure}

As illustrated in Figure~\ref{fig:policy_initialization}, the initialization process of LLMs consists of two primary phases: pre-training and instruction fine-tuning. During pre-training, models develop fundamental language understanding through self-supervised learning on large-scale web corpora~\citep{Sun2024MOSS,weber2024redpajamaopendatasettraining,liu2024datasetslargelanguagemodels}, following established power-law relationships between computational resources and performance~\citep{kaplan2020scaling,hoffmann2022trainingcomputeoptimallargelanguage}. Instruction fine-tuning then transforms LLMs from simple next-token prediction to generating human-aligned responses~\citep{FLAN, Scaling_Flan}. For models like o1, incorporating human-like reasoning behaviors is crucial to enable more sophisticated exploration of solution spaces. We summarize six key behaviors that can be activated through prompts or learned through expert trajectories distillation from LLMs.

\subsection{Pre-Training}
\label{sec:pre-training}
Pre-training establishes basic language understanding and reasoning capabilities in LLMs through exposure to massive text corpora~\citep{Radford2018ImprovingLU,Analysis_Reasoning_Corpus}. For o1-like models, these core competencies serve as the basis for advanced behaviors developed through subsequent learning and search.

\subsubsection{Language Understanding and Generation}
\label{subsubsec:language_understanding}
Pre-training cultivates diverse language capabilities through extensive exposure to natural language~\citep{Radford2018ImprovingLU}. At the syntactic level, models learn grammatical structures ranging from basic word order patterns to complex dependencies~\citep{manning2022human}. This syntactic foundation enables pragmatic understanding, including discourse markers and contextual language use, allowing models to adapt different styles across tasks~\citep{LLMChatbot}. Generation capabilities progress from basic grammatical coherence to sophisticated features like long-range consistency and complex narrative structures~\citep{tian2024narratives}. Through multilingual training data, models develop cross-lingual abilities, enabling zero-shot transfer and cultural understanding across languages~\citep{bloom,alves2024tower}. Research shows language understanding emerges hierarchically: syntactic patterns appear early, while logical consistency and abstract reasoning develop later, suggesting the importance of training duration and data composition beyond model scaling~\citep{Hierarchy_Encoders,Learning_Syntax}.

\subsubsection{World Knowledge Acquisition and Storage}
\label{subsubsec:world_knowledge}
Pre-training enables comprehensive knowledge acquisition across factual, procedural, and conceptual domains through diverse corpora processing~\citep{Radford2019LanguageMA,brown2020languagemodelsfewshotlearners}. Models develop rich semantic networks of factual knowledge from encyclopedic sources and academic literature, enabling cross-domain reasoning and novel insights~\citep{Acquire_Factual_Knowledge}. Domain expertise emerges from specialized technical content, manifesting in advanced capabilities like mathematical proofs and scientific analysis~\citep{shao2024deepseekmathpushinglimitsmathematical, Qwen2.5-Math}. Procedural knowledge develops through exposure to instructional content and programming languages, enhancing systematic problem-solving abilities~\citep{ruis2024proceduralknowledgepretrainingdrives, PlaSma,Teach_Themselves_Program}. Mathematical and logical foundations form through formal mathematical texts, establishing logical reasoning capabilities~\citep{reasoning_survey,LLM_Mathematical_Reasoning}. Recent studies demonstrate that knowledge storage exhibits efficient compression and generalization properties~\citep{knowledge_compression}, with abstract concepts requiring more extensive training compared to factual knowledge~\citep{Physics_of_LLM_3.1}.

\subsubsection{Basic Reasoning Abilities}
\label{subsubsection:basic_reasoning}
Pre-training develops foundational reasoning capabilities through diverse reasoning patterns, emerging hierarchically from simple inference to complex reasoning. Pattern matching and analogical reasoning emerge as primary mechanisms, enabling generalization across domains~\citep{Analogical_Reasoning,Analogical_Reasoners,Thought_Propagation}. Logical inference abilities develop through exposure to massive code and mathematical proofs~\citep{Qwen2.5-Coder,Qwen2.5-Math}, while sequential processing capabilities emerge from procedural texts and mathematical derivations~\citep{lewkowycz2022solving}. These abilities enable complex problem decomposition and logical coherence.

\subsection{Instruction Fine-Tuning}
\label{subsubsection:instruction_fine_tuning}

Instruction fine-tuning transforms pre-trained language models into task-oriented agents through specialized training on instruction-response pairs across diverse domains~\citep{Survey_Instruction_Tuning,Instruction_Pre_Training}. This process alters the model's behavior from pure next-token prediction to purposeful behavior~\citep{analysis_instruction_tuning,Layer_by_Layer}. The effectiveness of instruction fine-tuning depends primarily on two key factors: the diversity of the instruction dataset~\citep{T0} and the quality of instruction-response pairs~\citep{From_Quantity_to_Quality,CoachLM}. Research efforts have expanded both dimensions significantly, with \citet{Super-NaturalInstructions} developing a comprehensive dataset encompassing over \textit{1,600} distinct NLP tasks. FLAN~\citep{FLAN} demonstrated that models fine-tuned on high-quality instruction data can effectively generalize to novel tasks. \citet{Scaling_Flan} enhanced this capability through scaling across task count, model size, and incorporation of step-by-step reasoning protocols. Smaller-scale models like Alpaca~\citep{alpaca} achieved remarkable instruction-following capabilities through carefully curated high-quality training data. Self-Instruct~\citep{Self-Instruct} introduced methods for automated generation of instruction-response pairs using LLMs themselves. \citet{CoI} further established that fine-tuning on complex, multi-step instructions significantly enhances model capabilities and generalization.

\subsection{Human-like Reasoning Behaviours}
\label{subsubsection:human_like_reasoning}
\begin{table*}[ht]
\centering
\footnotesize
\begin{tabular}{p{3cm}p{8cm}p{2cm}}
\toprule
Reasoning Behavior & Exemplar & Source \\ 
\midrule
Probelm Analysis & So the user is requesting a bash script that\ldots Let's first understand the input and output formats\ldots So the requested output is '[1,3,5],[2,4,6]' & Coding \\
\rowcolor{Gray}
Task Decomposition & Implementation Steps: 1. Capture input string as argument.\newline 2. Remove any spaces (if any). 3. Parse the input string\ldots & Coding \\
Task Completion & Let me try coding the bash script step by step. \newline First, let's write the skeleton: & Coding \\
\rowcolor{Gray}
Alternative Proposal & Option 1: Take the odd positions: \newline Option 2: Try mapping as per an assigned code: & Cipher \\
Self-Evaluation & Check the number of letters.\ldots Let's test this mapping.\ldots Let's check this with the second pair.& Cipher \\
\rowcolor{Gray}
Self-Correction & Wait, the correct formula is: $\mathrm{pH} = 7 + 0.5 \times \log \left(\frac{K_b \, \text{for base}}{K_a \, \text{for acid}} \right)$ & Science \\
\bottomrule
\end{tabular}
\caption{Exemplars of Human-like Reasoning Behaviors from o1 blog~\citep{o1_blog}}
\label{tab:reasoning-behaviors}
\end{table*}

While instruction-tuned models demonstrate general task competency and user intent understanding, models like o1 require a more sophisticated repertoire of human-like reasoning capabilities to fully leverage their potential. As shown in Table~\ref{tab:reasoning-behaviors}, similar to \citet{Reverse-o1}, our analysis of o1's behavior patterns identified six human-like reasoning behaviors that help o1 better explore the solution space. We examine the implementation of these reasoning behaviors through two complementary perspectives: supervised fine-tuning and prompt engineering \citep{thinking-calude}.

\paragraph{Problem Analysis}
Problem analysis serves as a crucial initialization process where the model reformulates and analyzes the problem before solving it~\citep{kondrakunta2018toward}. This process involves multiple steps: explicit problem restatement to verify understanding, identification of implicit constraints, and transformation of abstract requirements into concrete, actionable specifications. \citet{Proactive_Dialogues} advance this concept through Proactive Chain-of-Thought, where models actively analyze potential ambiguities before proceeding with problem-solving. In o1's blog on cipher solving, this manifests as careful observation of ciphertext patterns and explicit problem reformulation. As shown in Table~\ref{tab:reasoning-behaviors}, in coding tasks, it restructures inputs into matrices and precisely generates expected outputs. Problem analysis reduces ambiguity in problem interpretation, constructing a more advantageous initial state for the subsequent phase~\citep{Clarification}.

\paragraph{Task Decomposition}
When encountering complex problems, humans typically decompose them into several manageable subtasks~\citep{Least-to-Most}. As shown in Table~\ref{tab:reasoning-behaviors}, in coding tasks, o1 decomposes the problem into several subtasks, including capturing input strings, removing spaces, and parsing input strings. \citet{Complex_Tasks_Compositional} introduce Compositional Fine-Tuning (CFT), a technique that explicitly divides a target task into its constituent components. Recent studies show that models can effectively perform such decompositions when guided by carefully structured prompts~\citep{Decomposed_Prompting, Successive_Prompting}. Importantly, the decomposition process is adaptive and context-aware, with the model dynamically adjusting the granularity and structure of subtasks based on the problem's complexity and uncertainty levels~\citep{ADaPT,Adaptive_Solver}.

\paragraph{Task Completion}
Following problem analysis and task decomposition, the model generates solutions through step-by-step reasoning based on clarified problems and decomposed subtasks~\citep{Reasoning_with_LLMs,reasoning_survey}. This behavior forms the foundation for all other reasoning processes, where successful solutions lead to subsequent subtask processing, while problematic solutions trigger the generation of alternatives or self-correction behaviors. Step-by-step generation significantly enhances models' complex reasoning capabilities~\citep{wei2022emergentabilitieslargelanguage,CoT_survey}. For LLMs, this ability can be activated through prompts containing reasoning processes~\citep{wei2022chain}, or even through simple instructions like ``Let's think step by step''~\citep{Zero_shot_CoT}. Smaller models can acquire this capability through distillation on extensive step-by-step reasoning data~\cite{shridhar2023distilling,Distilling_Step_by_Step}. Recent research indicates that sampling multiple solutions substantially increases the probability of generating correct answers~\citep{Large_Language_Monkeys}, while selecting final answers based on marginal probabilities effectively improves overall accuracy~\citep{SC}.

\paragraph{Alternative Proposal}
When faced with reasoning obstacles or dead ends, the ability to generate diverse alternative solutions becomes crucial~\citep{jintelligence9020023,Divergent_Thinking}. As shown in Table~\ref{tab:reasoning-behaviors}, o1 demonstrates this capability in cipher solving by systematically proposing several options. Divergent CoT~\citep{Fine_Tuning_Divergent_Chains} fine-tunes models to generate multiple solutions in a single inference, significantly improving performance on complex reasoning tasks. The generation of alternative proposals can be activated through prompting strategies. Progressive-Hint Prompting~\citep{php} leverages historical solution attempts to guide current reasoning, while Exchange-of-Thought~\citep{Exchange_of_Thought} enriches the solution space by incorporating insights from other models. This systematic exploration of alternatives not only expands the search space but also enables iterative refinement through solution comparison, leading to more well-reasoned outputs.

\paragraph{Self-Evaluation}
Following task completion, self-evaluation serves as a critical verification mechanism to validate the correctness of proposed solutions~\citep{Self-Verification}. As shown in Table~\ref{tab:reasoning-behaviors}, in o1's cipher example, the model compares plaintext and ciphertext letter by letter, expressing self-evaluation through explicit feedback such as ``Let's check'' or ``Let's test''. This evaluation capability can be enhanced through two primary approaches: implementing detailed evaluation criteria to instill self-evaluation abilities~\citep{AnthropicCAI,aor}, or utilizing self-debate for cross-validation~\citep{MAD,corex}. \citet{zhang2024self} introduces Self-Knowledge Tuning to strengthen models' self-evaluation capabilities, enabling more reliable assessment of their own reasoning processes. \citet{liu2024minds} demonstrates that distilling self-evaluation abilities into smaller models significantly improves their reasoning performance.

\paragraph{Self-Correction}
When encountering manageable errors in the reasoning process, models employ self-correction behaviors to address them~\citep{self_correction_survey}. In o1's demonstration, encountering signals like ``No'' or ``Wait'' triggers the correction process. As shown in Table~\ref{tab:reasoning-behaviors}, in o1's science example, the model identifies errors in formula generation and produces the correct formula through self-correction. \citet{LLM_cannot_self_correct} highlights the challenges of self-correction without external feedback, while \citet{Intrinsic_Self_Correction} demonstrates that unbiased prompts and zero temperature settings can unlock intrinsic correction capabilities. \citet{Learning_to_Check} introduces step-level analysis that significantly improves self-correction performance across multiple datasets.

While the behaviors described above provide insights into o1's human-like reasoning capabilities, they represent only a subset of its comprehensive reasoning framework. The model demonstrates sophisticated adaptive behaviors that extend beyond these fundamental patterns, dynamically adjusting its problem-solving strategies based on task-specific requirements and constraints. Through systematic analysis of these behaviors, we gain valuable insights into how o1 navigates complex textual domains and modulates its reasoning behaviors across diverse contextual environments.

\subsection{Speculation About the Policy Initialization of o1}
\label{subsection:speculation_policy_initialization}
During the progression from pre-training to instruction following, the model gradually constrains its action space. Policy initialization plays a critical role in developing o1-like models, as it establishes foundational capabilities that influence subsequent learning and search processes. The policy initialization phase encompasses three essential components: pre-training, instruction fine-tuning, and the development of human-like reasoning behaviors. While these reasoning behaviors are implicitly present in LLMs after instruction fine-tuning, their effective deployment requires activation through either supervised fine-tuning or carefully crafted prompts. Below, we outline several foundations for effectively utilizing these human-like reasoning behaviors.

\paragraph{Long-Text Generation Capabilities}
During inference, LLMs need to generate a large number of tokens to encompass complex and diverse reasoning behaviors, which requires sophisticated long-context modeling capabilities. While current LLMs have significantly improved in processing long texts~\citep{Survey_longcontext}, their capacity for generating lengthy content remains limited. \citet{LongWriter} introduces AgentWrite, an agent-based automated data construction pipeline. By fine-tuning on the constructed LongWriter-6k dataset, the approach substantially enhances LLMs' long-text generation capabilities. Similarly, \citet{Self-Lengthen} proposes Self-Lengthen, which iteratively fine-tunes on data constructed by the Extender, continuously improving LLMs' long-form text generation performance.

\paragraph{Shaping Human-like Reasoning Behaviors Logically}
Beyond generating extensive outputs, models must develop the capability to orchestrate human-like reasoning behaviors in a logically coherent manner. This orchestration requires sophisticated decision-making; for instance, when self-evaluation identifies an error, the model must strategically determine whether to engage in self-correction or explore alternative solutions. While these human-like reasoning behaviors enable comprehensive exploration of solution spaces, they simultaneously introduce computational complexity and demand enhanced logical reasoning capabilities. Current research demonstrates that exposure to programming code and structured logical data significantly strengthens a model's reasoning capabilities~\citep{sun2024survey,Impact_of_Code}. However, the systematic organization and sequencing of these human-like reasoning behaviors remains an open research challenge, particularly in determining optimal decision points for deploying specific reasoning behaviors.

\paragraph{Self-Reflection}
We identify behaviors such as self-evaluation, self-correction, and alternative proposal as manifestations of the model's self-reflection capability. Self-reflection addresses a fundamental limitation of autoregressive models: the inability to revise previously generated content~\citep{Self_refine}. Additionally, self-reflection demonstrates the model's self-knowledge~\citep{Say_IDK, yin2023large}, enabling it to spontaneously recognize flaws in its generated content. Current research indicates that this capability is not easily acquired and cannot be effectively learned through parameter-efficient fine-tuning methods~\citep{Physics_of_llm_2.2}.

\subsection{Challenges of Policy Initialization for Reproducing o1}
\label{subsection:challenges_in_policy_initialization}

While policy initialization establishes crucial foundations for o1-like models, several significant challenges emerge in implementing this approach effectively.

\paragraph{How to balance sampling-efficiency and sampling-diversity?}
Policy initialization faces a critical trade-off between sharping the action probability distributions for efficient sampling and maintaining sufficient diversity for exploration. While learning from human demonstrations helps constrain the action space, excessive convergence to fixed strategies can limit the discovery of superior approaches during search phases~\citep{GEM}. This challenge is evident in comparing AlphaGo and AlphaGo Zero, initialization from human data provides a strong starting point but may inadvertently restrict exploration of potentially better strategies.

\paragraph{How to ensure domain generalization of reasoning behaviors?}
Current research focuses on replicating o1's behaviors in specific domains such as mathematics~\citep{AlphaMath} and coding~\citep{o1-coder}. However, o1's behaviors are not limited to domain-specific reasoning behaviors. For instance, in safety tasks, models need to perform behaviors that verify whether generated content complies with safety guidelines. Therefore, during the policy initialization process, models should not be restricted to domain-specific reasoning behaviors. Since it is impractical to specify corresponding reasoning behaviors for all tasks, designing reasoning behaviors with strong domain generalization capabilities becomes crucial.       

\section{Reward Design}
\label{sec:reward_modeling}
In reinforcement learning, the agent receives feedback from the environment in the form of a reward signal and seeks to maximize its long-term reward by improving its policy. The reward function, denoted as \( r(s_t, a_t) \), represents the reward associated with the agent's action \( a_t \) in state \( s_t \) at time step \( t \). The reward signal is crucial in guiding both the training and inference processes, as it defines the desired behavior of the agent through numerical scoring. While the same optimal policy can be learned from various reward designs \citep{PotentialFunc}, a well-designed reward signal can accelerate both the convergence of learning and the efficiency of the search process.

\begin{figure}[h]
    \centering
    \includegraphics[width=\linewidth]{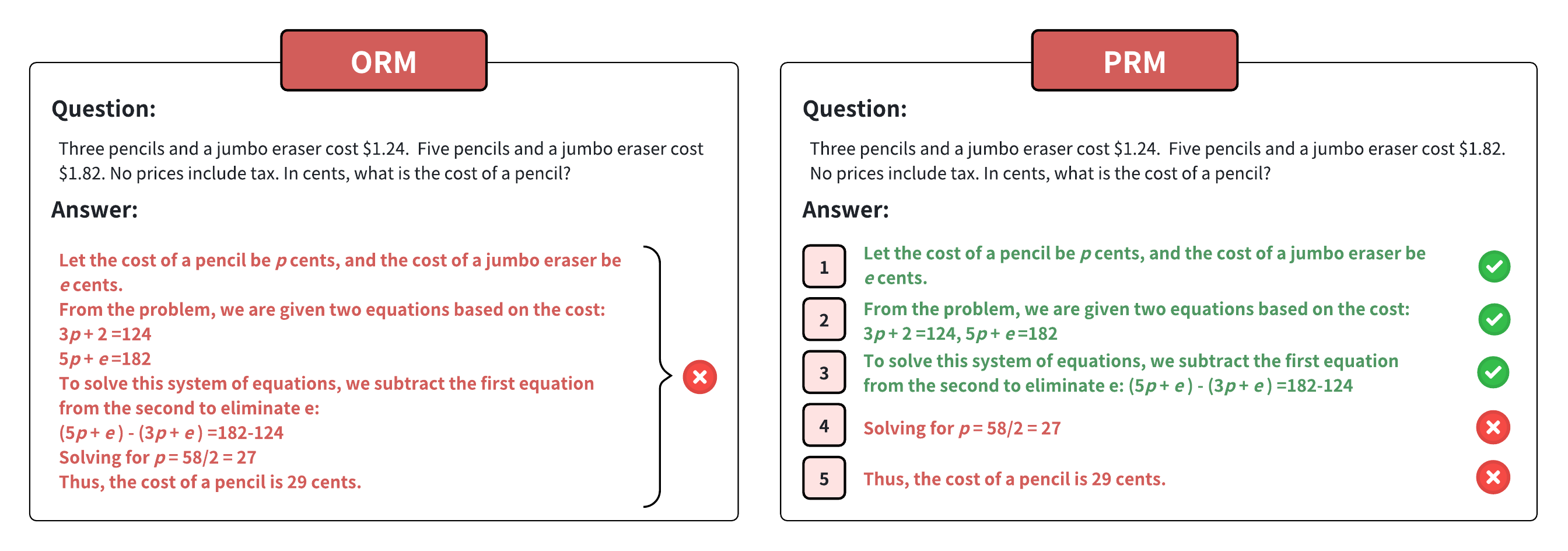}
    \vspace{-2em}
    \caption{Outcome reward vs Process reward. The figure shows two different types of reward: outcome reward (left) and process reward (right). The entire solution is incorrect due to the penultimate line. Therefore, for outcome rewards, the whole solution is marked as incorrect, while for process reward, the previous three steps are marked as correct and the last two steps as incorrect.}
    \label{fig:reward_modeling}
    \vspace{-1em}
\end{figure}

This section provides an overview of the current reward design methods for large language models (LLMs). We begin by comparing the \textbf{outcome reward} and the \textbf{process reward}. Next, we introduce reward design methods that have been specifically proposed for LLMs. We also review reward design techniques that are widely used in reinforcement learning but have not yet been applied to LLMs. Finally, based on our comparison of these different methods, we offer a speculative discussion on the reward design for o1.

\subsection{Outcome Reward vs Process Reward}
The outcome reward involves assigning a score based on whether the outputs of a large language model (LLM) meet predefined expectations ~\citep{OpenAIMathVerifierORM, shao2024deepseekmathpushinglimitsmathematical}. For instance, \citet{OpenAIMathVerifierORM} uses outcome reward to assess whether the solution provided by the LLM to a mathematical problem is correct. Since ground truth is generally available for tasks such as solving mathematical problems, outcome reward is often straightforward.

Although the outcome reward is relatively easy to construct, it lacks supervision for intermediate steps. For instance, a solution to a mathematical problem that leads to the correct answer may still contain errors in the intermediate steps \citep{VerifySbyS}. As a result, using the outcome reward may cause the LLM to generate incorrect solution steps, which could negatively affect performance. Moreover, outcome reward is sparse, as it does not reward intermediate steps, making it challenging to learn step-level policies. Despite these limitations, it remains widely used in reinforcement learning for LLMs due to its simplicity.

In contrast to the outcome reward, the process reward provides a reward signal not only for the final step but also for intermediate steps \citep{VerifySbyS, yin2024reasoning}. For example, \citet{VerifySbyS} involves human annotators rewarding the intermediate steps in a mathematical solution generated by an LLM. The outcome reward can be seen as a special case of the process reward, where the rewards for intermediate steps are all set to zero. 

Depending on the granularity of the actions, the process can be classified into token-level \citep{FromRtoQ} and step-level \citep{VerifySbyS}. The step-level segmentation is flexible. \citet{SemanticUncertainty} uses information entropy to guide step segmentation while methods in mathematical problem-solving typically use newline characters as the split symbol \citep{VerifySbyS}.

Although process rewards show promise, they are more challenging to learn than outcome rewards. For instance, the reward design in \citet{VerifySbyS} depends on human annotators, making it costly and difficult to scale. There are some automatic ways to transform outcome rewards to process rewards, which is called reward shaping in traditional reinforcement learning.
We introduce the methods of reward shaping in Section \ref{subsubsec:reward_shapping}.

\subsection{Reward Design Methods}
\label{subsec:reward_design_for_LLMs}
Since outcome reward can be seen as a special case of process reward, many reward design methods can be applied to the modeling of both outcome reward and process reward. The resulting models are typically referred to as the Outcome Reward Model(ORM) and the Process Reward Model(PRM). We review these reward design methods for LLMs, categorizing them based on whether or not they have direct access to the reward signal from the environment.

\subsubsection{Reward from Environment} 
The most straightforward approach to designing a reward is to directly utilize the reward signal from the environment, or learn a model to simulate the reward signal from the environment.

\paragraph{From Realistic Environment} 
Many environments can provide effective reward signals, for example, code generation can receive reward signals from a compiler or interpreter. \citet{StepCoder,ExcutionCodeGen,AutoIF} have shown that the quality of code generation is improved with compiler feedback. \citet{PLUM} generates test cases for input questions and uses the testing outcomes as the reward signal. ~\citep{nl_to_code_translation_with_execution} measures the similarity of different programs based on their execution results. \citet{OpenCodeInterpreter} uses the execution results of a program as guidance for refining a code generation model. For some real-world tasks, the environment is just a sandbox. In web shopping \citep{AgentBench}, the environment is represented by a real HTML page, and the reward is calculated based on the score the LLM obtains through clicks and searches. In more general scenarios, sandboxes are created to allow the policy model to interact directly with, such as in environments like MineDojo \citep{MineDojo} or Alfworld \citep{Alfworld}.

\paragraph{From Simulating Environment} 
While some environments can provide valid feedback, interacting with them to obtain the reward signal can be costly, or the feedback may not be available at test time. For example, during testing, we might not have test cases to validate whether the program generated by the LLM is correct. In such cases, a reward model is required to simulate the reward signal from the environment. For instance, \citet{VerifySbyS} trains a verifier to predict whether the model's solution to a mathematical problem is correct at test time.

Simulating the reward signal enables the LLM to obtain feedback at any time. However, using such a reward model during learning or search can lead to the problem of distribution shift. As the policy is updated during learning and search, the reward model, which is trained on data from interactions between the old policy and the environment, may fail to adapt to the new policy. This issue, known as reward optimization, has been discussed in \citet{RewardOverOptim, inverse_inference_time_scaling}. Thus, the reward model must be updated in tandem with the policy improvements. In more general domains, simulating the environment aligns with the concept of a world model, where the transition probabilities of states must be further simulated, allowing for a more accurate reward model. We discuss the world model in Section \ref{sec:WorldModel}.

\paragraph{From AI Judgment}
AI Judgment involves using general AI assistants to provide the reward signal, serving as an alternative to relying on realistic environments, thereby avoiding the high costs associated with environment construction and interaction. For instance, it is common to use a powerful LLM, such as GPT-4, to assess the performance of an AI assistant \citep{MT-Bench}. While AI judgment can be viewed as a form of reward model, it does not face the issue of reward optimization, as it is not dependent on the policy model. As a result, AI judgment remains effective even when the policy model undergoes updates. The LLM is an implementation of the world model, which also highlights the effectiveness of building a world model to provide reward signals.

\subsubsection{Reward Modeling from Data}
For some environment, the reward signal from environment is unavailable, which can not be simulated. For example, it is difficult to judge whether the response from an AI assistant is good or not. Fortunately, it is easier to collect the expert data or the preference data than giving a reward. With the expert data or preference data, we can also learn a model to provide effective rewards.

Learning rewards from preference data is widely recognized in the LLM community, particularly due to the success of RLHF \citep{rlhf-anthropic}. However, learning rewards from expert data, also known as Inverse Reinforcement Learning, has not been widely used for LLMs, which may represent a promising technique for o1.

\paragraph{Learning Rewards from Preference Data}
Preference data is collected by ranking multiple responses from LLMs to the same question. Using the Bradley-Terry model, an outcome reward can be derived based on the pair-wise comparison \citep{Bradley1952RankAO}. Learning rewards from preference data for LLM is first proposed in \citet{DeepRLFromHF}, which introduced the preference signal to tackle complex reinforcement tasks without access to a reward function, achieving significant success in Atari games and robotics. They further developed this approach for text summarization~\citep{learn_to_summarize_with_human_feedback}.
This technique is then used for the development of ChatGPT, where preference signals are used for aligning policy behavior with human values ~\citep{AnthropicCAI}. 

Learning from preference data has been widely used in alignment of LLMs. However, it is crucial to construct the preference data that accurately reflects the actual performance of downstream tasks. For example, \citet{MisleadRLHF} show that using human preferences as supervision may degrade the true performance of the model.

\paragraph{Learning Rewards from Expert Data}
Inverse Reinforcement Learning (IRL) is a method for learning rewards from expert data, with the goal of recovering the reward function that the expert is optimizing. This is achieved by fitting the reward function to the trajectories generated by the expert and maximizing the recovered reward ~\citep{ApprenticeLearn, IRL}. Many IRL approaches integrate adversarial learning techniques. \citet{IQLearn} introduces the convex conjugate function to avoid the need for adversarial training. Compared with learning reward from preferences, the data for  IRL is easier to be collected. However, IRL typically involve adversarial training, which makes the learning more complicated than learning rewards from preferences. Despite that IRL is well-known in reinforcement learning, there is no empirical evidence that it has been used for large-scale reinforcement learning for LLM.

\subsubsection{Reward Shaping}
\label{subsubsec:reward_shapping}
The reward signal from some environments may not be effective; for example, it could be an outcome reward rather than a process reward. In such cases, the reward can be reshaped to make it denser and more informative, a process known as reward shaping. For instance, \citet{SCoRE} attempts to train an LLM to self-correct via reinforcement learning and discovers that proper reward shaping can prevent learning collapse. ~\cite{MathShepherd} reshaped the outcome reward by estimating the value function using Monte Carlo sampling, employing the Q-value $Q_\pi(s_t, a_t)$ as the reward during intermediate steps. \textbf{However, since the value function depends on the policy $\pi$, the value function estimate from one policy may not be a valid reward function for another policy.} This is empirically validated by \citet{xiong2024rlhflowmath}, who found that using a value function from another policy for reward shaping is harmful.

\citet{PotentialFunc} defines potential-based reward shaping and proves that two reward functions can lead to the same optimal policy if and only if they satisfy the following equation:
\begin{equation}
F(s_t, a_t) = r(s_t, a_t) + \gamma \phi(s_{t+1}) - \phi(s_t),
\label{eq:reward-shaping}
\end{equation}
where $F(s_t, a_t)$ and $r(s_t, a_t)$ are two reward functions that yield the same optimal policy, $\gamma$ is the discount factor, and $\phi$ is a potential function. Potential-based reward shaping suggests that the reward function $r(s_t, a_t)$ can be reshaped without altering the optimal policy, as long as the shaping satisfies Equation \ref{eq:reward-shaping}. \citet{DeepMindAdvReward} uses the theory of potential-based reward shaping to transform outcome rewards into process rewards.

The reward learned from preference data can also be reshaped. For example, \citet{dpo}, \citet{FromRtoQ} and \citet{DPOMeetPPO} demonstrate that DPO implicitly makes potential-based reward shaping \citep{PotentialFunc} on the reward learned from preference data.

While reward shaping can be advantageous, it may also be harmful. Improper reward shaping can negatively affect both learning and search processes. As a result, reward shaping requires careful design and often entails incorporating inductive bias.

\subsection{Speculation About the Reward Design of o1}

We now turn to discuss our assumptions regarding the reward design for o1. Given that o1 could handle multi-task reasoning, its reward model is likely to incorporate multiple reward design methods. For complex reasoning tasks, such as mathematics and code, where responses typically involve long chains of reasoning, a process-reward model is more likely to be employed to supervise the intermediate process, rather than ORM. Techniques like reward shaping can help derive process rewards from outcome rewards.

When no reward signal is available from the environment, we suspect that o1 will need to rely on learning from preference data \citep{DeepRLFromHF} or expert data \citep{IQLearn}. 

Given that o1 can be fine-tuned using few-shot examples, we suspect that it has a robust reward model trained on a large and diverse dataset spanning a wide range of domains. It can be adapted to a new domain easily through ground truth and solution pairs. Moreover, it is more likely to predict rewards by generating with LLM rather than through value heads.

\subsection{Challenges of Reward Design for Reproducing o1}
\label{sec:challenge_of_reward_design}

\paragraph{How to overcome distribution shift?} The reward model learns from an existing dataset distribution and the out-of-distribution problem still remains significant today~\citep{RewardOverOptim}, especially as the LLM continues to explore and learn from feedback. The reward provided by the proxy model deviates from the golden reward when the distribution of the policy model changes, as the trajectory becomes unseen during the training of the reward model and heavily depends on its generalization. Scaling the parameters of the reward model and increasing the amount of data can mitigate this issue but does not completely solve it. Iteratively training the reward model provides a more direct solution, but it still requires human involvement in the loop.

\paragraph{How to design fine-grained reward for language model?}
Unlike in Atari games or robotic environments, language presents a unique challenge because the definition of a step or action can vary in granularity: token-level, step-level or solution-level. In many situations, such as fitting the model to human preferences, it is more natural to judge the entire solution rather than each token individually, as higher-order semantics in language emerge from the combination of tokens. However, using token combinations as actions results in an action space that is too large to define or learn a reward function for. The potential action space grows exponentially, leading to a long tail of actions. Meanwhile, the sparsity of the reward signal increases with the length of each step as we mentioned before.

\paragraph{How to choose data when modeling reward for complex tasks?}
As the complexity of tasks increases, choosing the appropriate type of feedback becomes increasingly challenging. Recent studies have shown that using preference-based feedback for tasks like code generation or mathematical reasoning may actually degrade the performance of the policy model \citep{MisleadRLHF}. Additionally, the question of how much data is necessary to accurately capture the intended behavior remains underexplored. The difficulty of evaluating whether a reward is effective also grows as the complexity of the task increases.

\subsection{Generalization}
The previous section primarily focused on reward design for specific tasks, such as math and coding. When addressing more general tasks, we need to create a broader environment. Practically, according to OpenAI's 5-stages plan for AGI, o1 has been a strong reasoner, the next stage is to train an agent that can interact with the world and address real-world problems. To achieve this goal, we need a reward model to provide the reward signal for the agent taking actions in a realistic environment. In this section, we focus on how to construct a general reward signal, which can be divided into two components: the reward ensemble and the world model.
\subsubsection{Reward Ensemble}
An intuitive way to build a reward signal for a general task is through an ensemble rewards in specfic domains. \citet{DMoERM} trains the reward model as a form of MoE \citep{jiang2024mixtral,turn-waste-into-worth}, where each expert is trained on a different task to provide the corresponding reward signal, and the outputs are aggregated finally. \citet{LASER} frames this problem as a multi-armed bandit question, learning to select the most suitable reward model. In short, the crucial question is for these methods is how to combine the reward signals effectively.

\subsubsection{One for all: World Model}
\label{sec:WorldModel}
The world model can not only provide the reward signal but also predict the next state~\citep{world_models,NavigationWorldModel}. Some works claim that a video generator is a world model since the video generator can predict the images in the future time steps \citep{RingAttention, NavigationWorldModel, WorldDreamer, Sora}. \citet{WorldModel} also proposes a framework where world model do not need to predict the next state, instead they predict the representation of the next state, which is easier and more efficient than predicting a image. This is inline with Muzero \citep{Muzero}, which also let the environment simulator to predict the representation of the next state in the game of go. The current works of world model focus on modeling of the next state prediction, but we believe modeling the reward signal is also critical and challenging for an agent to accomplish real-environment tasks.

\section{Search}
\label{search}

\begin{figure}[t]
    \centering
    \includegraphics[width=\linewidth]{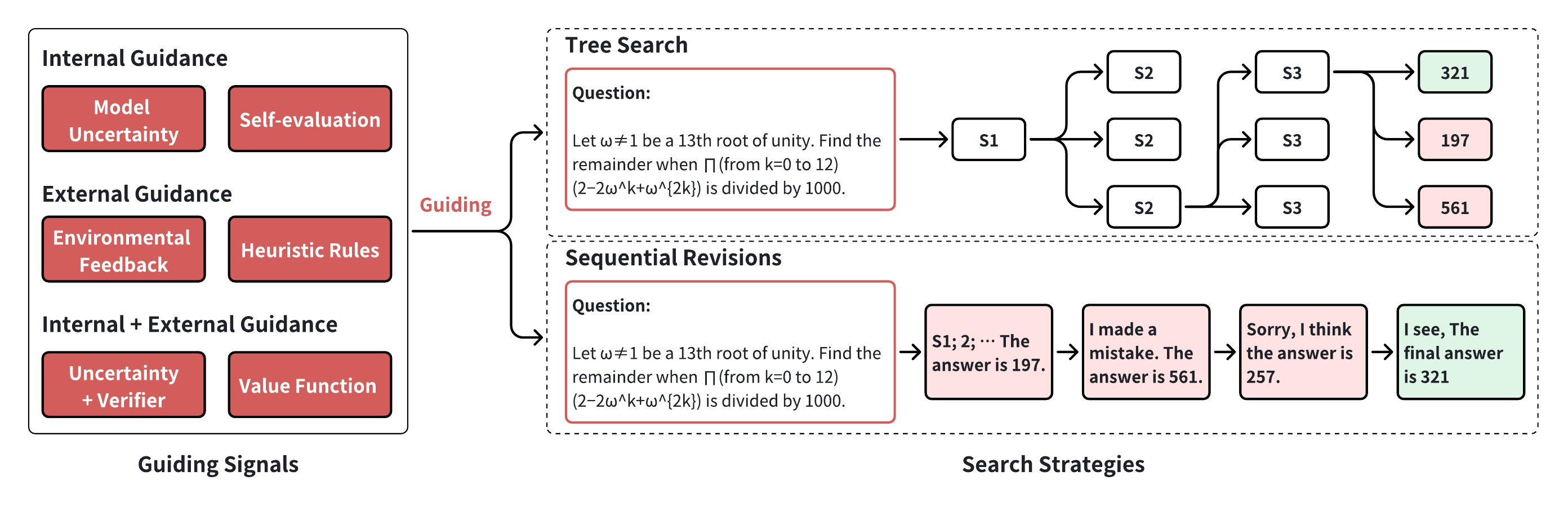}
    \vspace{-1em}
    \caption{The two key aspects of search are the guiding signals for solutions selection and the search strategies to get candidate solutions. We divide the guidance used during the search process into two types: internal guidance and external guidance. We also divide search strategies into two types: tree search and sequential revisions. Search strategies are used to obtain candidate solutions or actions, while guiding signals are used to make selections on candidate solutions or actions.
    S1 means step 1.}
    \vspace{-1em}
\label{fig:search_overview}
\end{figure}

For LLMs, performing random sampling during generation has become a mainstream method to improve output quality, with techniques such as nucleus sampling being prominent examples~\citep{top_p}.
Furthermore, many studies~\citep{SPoC, CodeX} have observed that the pass@k metric improves consistently as the number of model samples increases.
\citet{Large_Language_Monkeys} show that even small models can outperform large models leveraging search.
This suggests that language models have the potential to explore the correct solution by more sampling during inference, which requires consuming more inference-time computations.
Search refers to the process of finding the correct solution through multiple attempts or strategic exploration based on certain guidance, such as rewards or heuristic rules.
Well-known inference strategies like self-consistency ~\citep{SC} and Best-of-N (BoN)~\citep{OpenAIMathVerifierORM} both can be seen as search methods.
For model like o1, which are designed to solve complex reasoning tasks, search may play an important role in both training and inference processes.
In this section, we first analyze the possible role of search in o1, and then introduce the details of various promising search methods which have the potential to be used in the construction of o1-like models.

\subsection{The role of Search in o1}

Search often relies on guiding signals thus can be viewed as a process of strategy iteration, we call it search policy.
Compared to naive sampling, search policies are usually more likely to find better solutions.
Solutions generated by search policies can either be directly used as the final output or incorporated into training algorithms, like Expert Iteration~\citep{expert-iteration}, to iteratively improve the policy.
We believe that \textbf{search plays a crucial role in both the training and inference processes of o1}.
We will refer to the search during these two stages as \textit{training-time search} and \textit{test-time search}, respectively.

In the training phase, the trial-and-error process in online reinforcement learning can also be seen as a search process~\citep{Sutton_book}, where the agent performs naive sampling based on its own policy and learns solution that yield high rewards.
However, since o1 involves longer inference lengths and includes human-like reasoning behaviors, the search space becomes large, and naive sampling may become inefficient. 
Therefore, some advanced search strategies are needed to more efficiently explore better solutions and use them as training data to update the policy model.
This process can be iteratively conducted during training.
In the inference phase, o1 demonstrates that increasing compute during inference with more time spent thinking can continuously improve model performance~\citep{o1_blog}.
We argue that o1's way to think more can be seen as a kind of search, using more inference-time computation to find better answers.

The two key aspects of search are the guiding signals for the search and the search strategies to get candidate solutions.
Search strategies are used to obtain candidate solutions or actions, while guiding signals are used to make selections.
We first discuss the guiding signals for the search process, dividing them into internal and external guidance.
Inspired by \citet{Scaling_test_time_compute}, we categorize the search strategies into tree search and sequential revisions.
It is worth noting that these two classification dimensions are orthogonal, for example, tree search methods can utilize either internal or external guiding signals.
We present a schematic diagram of these categories in Figure \ref{fig:search_overview}.

\subsection{Search Guidance}
In the following sections, we will discuss methods for providing guiding signals to the search process.
Search based on internal guidance does not rely on real-world feedback from the external environment or proxy models, but instead uses certain states or evaluation capabilities of the model itself to guide the search process.
Classic text generation decoding algorithms, such as greedy decoding and beam search, typically use the probability of tokens or sequences as internal guidance for the search process.
External guidance is generally independent of specific policies and relies solely on environment- or task-related signals to guide the search process.

\subsubsection{Internal Guidance}

\paragraph{Model Uncertainty}
Model uncertainty~\citep{Predict_model_uncertainty} is a common form of internal guidance. 
Many studies leverage it to select high-quality responses from candidates, with self-consistency~\citep{SC} being a notable example.
Self-consistency uses majority voting or weighted sums to choose the answer with the lowest uncertainty. 
Universal self-consistency~\citep{Universal_SC} extends this approach to free-form responses, enabling large language models to select the most consistent answer without task-specific constraints, thus broadening its applicability.
Building on the idea that different sentences can express the same meaning, \citet{SemanticUncertainty} shift the focus to semantic rather than syntactic uncertainty.
They use NLI models~\citep{MNLI} for semantic clustering, treating two sentences as equivalent if they exhibit bi-directional entailment.
The entropy of the semantic distribution is then used to measure uncertainty.
This semantic entropy has also been applied to detect hallucinations~\citep{semantic_entropy_for_hallucinations}, showing strong performance.
Model uncertainty is often derived through unsupervised methods, making it easy to be obtained.
However, its reliability depends heavily on the model's calibration~\citep{calibration_of_modern_neural_networks}.
\textbf{A well-calibrated model is essential for effectively utilizing model uncertainty.}

\paragraph{Self-evaluation}
Model uncertainty is a useful guide but does not directly reflect response accuracy.
For example, high-uncertainty answers can still be correct.
To improve model performance, self-evaluation aims to let models assess their own outputs, leveraging the assumption that evaluation is easier than generation, known as the generator-discriminator gap (DG-gap)~\citep{janleike2022why_excited_about_AI_assisted_human_feedback, network_of_network}.
A key application of the DG-gap is reinforcement learning from human feedback (RLHF)~\citep{learn_to_summarize_with_human_feedback}.
Compared to directly annotating the correct model output, researchers have found that human annotators' preferences for different candidate outputs can serve as effective and scalable supervisions.
This approach helps align the model's behavior with human intentions and values.
For instruction-following language models, LLM-as-a-Judge~\citep{llm_as_a_judge} is an efficient self-evaluation method, achieving high agreement with human evaluators on MT-Bench~\citep{MT-Bench}.
Using this approach,\citet{Self_rewarding} demonstrate that models can improve self-rewarding through iterative DPO training. 
Similarly,\citet{Meta_Rewarding} introduce LLM-as-a-Meta-Judge to enhance self-evaluation.
\textbf{The main challenge in task-specific self-evaluation is determining whether a DG-gap exists and how to fully exploit it for greater accuracy.}
It is important to note that some studies~\citep{LLM_cannot_self_correct} suggest that models cannot accurately evaluate answers without feedback.
To address this, we may need to scale up the model size, train it for self-evaluation, specify more detailed evaluation criteria.

\subsubsection{External Guidance}

\paragraph{Environmental Feedback}
Using environmental feedback corresponding to downstream tasks is one of the most commonly used forms of external guidance and is typically directly related to the evaluation metrics of downstream tasks.
Reward is a typical form of environmental feedback and is often used to guide the search process, directly corresponding to the final performance.
We have discussed details of reward design in Section \ref{subsec:reward_design_for_LLMs}.
During the search process, we may need multi-scale rewards for guidance, such as outcome rewards and process rewards.
Using rewards as guiding signals often requires constructing the environment or utilizing proxy feedback, which can improve search effectiveness but may also introduce additional computational overhead.
Moreover, when using proxy feedback such as a reward model for search, if the sampled solutions deviate significantly from the distribution of the reward model's training data, the search may actually lead to a decrease in performance.
Besides, environmental feedback that is indirectly related to the final performance can also be used to guide the search process, such as code compilation outcomes, correctness of intermediate steps in solving mathematical problems, unit tests, and more.
This type of feedback often serves as an alternative to feedback on the final task outcomes.

\paragraph{Heuristic Rules}
Many search algorithms use heuristics as guiding signals alongside environmental feedback.
Traditional informed search methods, such as greedy search and $A^*$ search, require the use of heuristic rules developed based on the specific information of the task to guide the search process~\citep{ai-book}.
Additionally, studies for enhancing LLMs' reasoning abilities~\citep{RAP, Tree_search_for_language_model_agents} also apply task-specific heuristics to guide search.
When we want to reduce the cost of environmental feedback or when environmental feedback is unavailable, using heuristic rules is a good alternative.

\subsubsection{Comparison of Internal and External Guidance}
This section compares internal and external guidance.
Internal guidance relies solely on the model, avoiding the need for external environments or ground truth, and is generally task-agnostic. 
Thus, \textbf{internal guidance is highly transferable and useful when specific evaluation criteria for downstream tasks are unavailable}.
External guidance relies on specific downstream task information, such as rewards from interactive environments or ground truth, making it more aligned with model performance and better at guiding search strategies.
However, it introduces construction costs and computational overhead.
During inference, the ground truth is often unavailable, and interacting with the environment or simulator is costly.
Additionally, external guidance from surrogate models (like a fixed reward model) can face out-of-distribution (OOD) issues (\ref{sec:challenge_of_reward_design}).
Therefore, external guidance during inference requires careful consideration.

\subsubsection{Combination of Internal and External Guidance}
Internal and external guidance can be combined to direct the search process, with typical approaches integrating the model's own uncertainty and proxy feedback from reward models.
For example, \citet{MathShepherd} and \citet{Scaling_test_time_compute} combine self-consistency with feedback from a process reward model to select the highest-quality responses.

The value function is another type of signal that combines both internal and external guidance.
In reinforcement learning, the value function estimates expected cumulative rewards from a state (V function) or state-action pair (Q function), guiding agents to select actions that maximize long-term rewards.
It typically relies on environmental reward signals and uses a separate neural network~\citep{OVM, ppo_mcts, MathShepherd}.
The calculation formula for the value function is as follows (refer to \citet{Sutton_book}):
\begin{equation}
v_{\pi}(s) \doteq \mathbb{E}_{\pi}[G_t \mid S_t = s] = \mathbb{E}_{\pi} \left[ \sum_{k=0}^{\infty} \gamma^k R_{t+k+1} \mid S_t = s \right], \, \text{for all } s \in \mathcal{S},
\label{eq:value_function}
\end{equation}

In this equation, \( v_{\pi}(s) \) denotes the value of state \( s \) under policy \( \pi \), while \( \mathbb{E}_{\pi} \) represents the expectation taken over all possible trajectories following the policy \( \pi \). The term \( G_t \) is the return, which is defined as the cumulative sum of future rewards discounted by the factor \( \gamma \), where \( \gamma \in [0, 1] \) determines the importance of future rewards relative to immediate rewards. Specifically, \( R_{t+k+1} \) is the reward received at time step \( t+k+1 \), conditioned on the state at time \( t \) being \( s \). Finally, \( \mathcal{S} \) refers to the set of all possible states in the environment.

The value function plays a fundamental role in reinforcement learning, as it quantifies the long-term expected reward for starting in a specific state and taking actions according to the given policy. By evaluating the value of states, the value function enables agents to compare and make informed decisions about which states or actions are more favorable under a particular policy.

\citet{FromRtoQ} show that DPO is an inverse Q-learning algorithm, where DPO model logits act as the Q function, enabling search processes guided by the value function.
They also demonstrate performance gains from applying beam search to the DPO model.
Additionally, some work~\citep{AlphaZero, Alpha-zero-like, AlphaMath} incorporates a value head into the policy model, sharing a large-scale backbone.
\textbf{The main challenge in using value functions lies in accurately estimating them}, especially in tasks with sparse rewards or high-dimensional outputs, such as large language model generation, where inaccuracies can significantly impact performance.

\subsection{Search Strategies}
We categorize search strategies into two types: \textbf{tree search} and \textbf{sequential revisions}.
Tree search generates multiple answers simultaneously, acting as a global search that explores a broader range of solutions.
In contrast, sequential revisions refine each attempt based on previous ones, functioning as a local search that may offer higher efficiency~\citep{Scaling_test_time_compute}.

\subsubsection{Tree Search}
\label{sec:tree_search}

\begin{figure}[t]
    \centering
    \includegraphics[width=\linewidth]{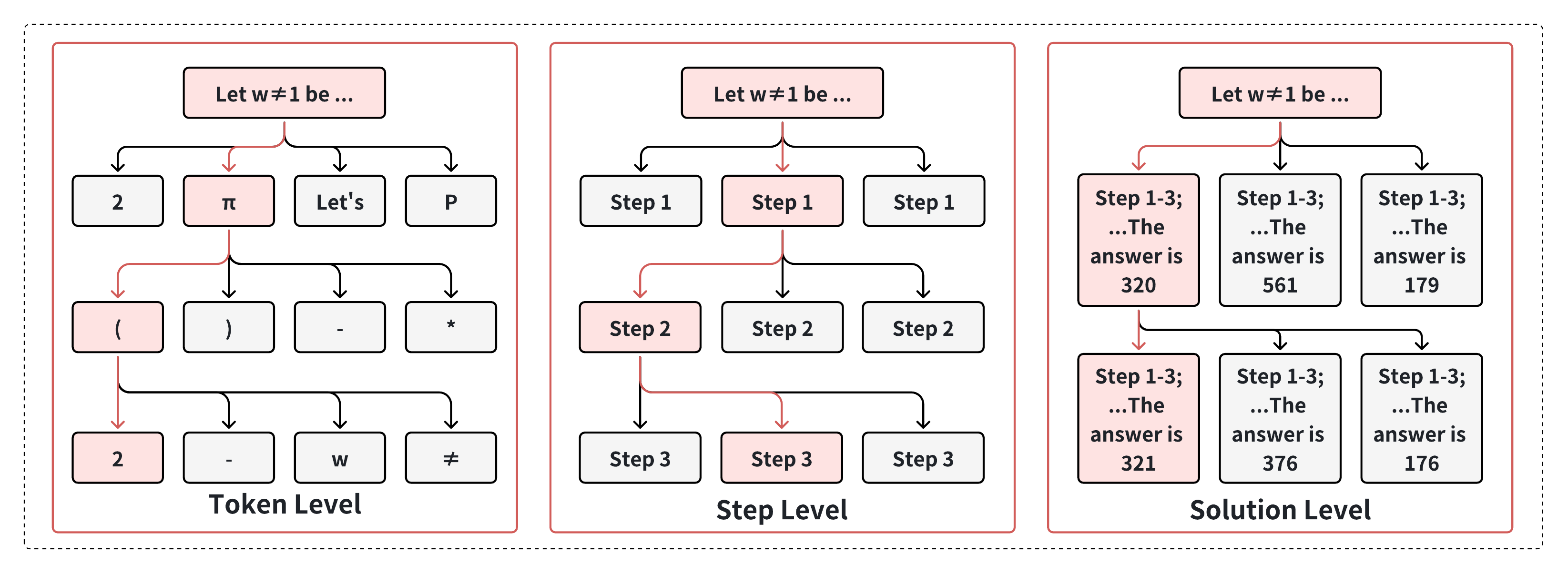}
    \caption{Definitions of search tree nodes at different granularities.
    The token level represents the finest granularity, while the solution level represents the coarsest granularity.}
    \label{fig:action_space_granularity}
\end{figure}

This section introduces tree search algorithms like Best-of-N (BoN), beam search, and MCTS.
BoN generates multiple independent candidate solutions but lacks dynamic adjustment of the model's probability distribution, leading to inefficiencies, such as over-sampling high-probability options. 
BoN can be seen as a special case of tree search with depth-1 nodes.
In contrast, other tree search strategies dynamically adjust at each step, balancing exploration and exploitation with heuristics, and can use lookahead search, backtracking, and pruning to improve efficiency and reduce sampling costs. 
An important question in using tree search algorithms on LLMs is defining the granularity of the tree node.
As shown in Figure \ref{fig:action_space_granularity}, common granularities of tree nodes include: token level, step level, and solution level.
The token level represents the finest granularity, while the solution level represents the coarsest granularity.
Generally, the smaller the granularity of the search tree nodes, the deeper the search tree becomes.

\paragraph{Best-of-n Sampling}
Best-of-N sampling (BoN) is a simple yet effective search method, which can be seen as a solution-level tree search.
It first generates multiple candidate solutions from the model, then selects the best solution via a reward model while discarding the rest~\citep{OpenAIMathVerifierORM}.
When BoN uses oracle rewards (e.g., comparing with ground truth), its accuracy improves with more samples, as coverage increases~\citep{Large_Language_Monkeys}.
However, in most cases, oracle rewards are unavailable, making that learning a reward model becomes the primary bottleneck.
Additionally, increasing the number of samples raises computational costs, further limiting BoN's scalability.

To reduce BoN's cost, \citet{Speculative_bon} propose speculative rejection, leveraging the correlation between partial and complete sequence scores. By scoring partial sequences and discarding low-scoring solutions early, it reduces computation while achieving BoN-like performance.
An alternative to multiple sampling is fine-tuning the policy model to mimic BoN’s distribution.
\citet{BoNBoN} theoretically prove that BoN is optimal under specific constraints and fine-tune the model using best-of-n and worst-of-n data with preference-based IPO loss and SFT loss.
Similarly, \citet{Variational_bon} introduce variational BoN (vBoN), which uses PPO to minimize the gap between the policy model and BoN's distribution, achieving superior performance without repeated sampling.
\citet{BOND} propose BOND, which fine-tunes the model to approximate BoN's distribution by minimizing Jeffreys divergence (a combination of forward and backward KL divergences).
Through iterative distillation, BOND reduces inference to a single sample while maintaining BoN-level performance.

\paragraph{Beam Search}
Beam search is a classic tree search algorithm that traditionally expands and prunes branches based on partial sequence probabilities.
Beam search is typically conducted at the token level.
Recent advancements in large language models have led to modifications of beam search by integrating additional guidance signals.
TreeBoN~\citep{TreeBoN} iteratively expands branches and prunes low-quality responses using token-level rewards from DPO~\citep{FromRtoQ}, maintaining high-quality sequences while reducing computation in a manner similar to beam search, on step level.
Similarly, \citet{OVM} train a value model from outcome supervision to score partial sequences, replacing token probabilities.
\citet{Self_evaluation_guided_beam_search} improve stochastic beam search~\citep{stochastic_beam_search}, balancing exploitation and exploration by using a policy model for self-evaluation for sampling instead of token probabilities.
\citet{Reward_step_by_step} propose a process reward model to guide greedy search, introducing backtracking when child nodes have negative rewards.
\citet{Scaling_test_time_compute} implement lookahead search via k-step rollouts for evaluating partial sequences, which can be seen as a specia case of MCTS.
Reward-guided beam search, compared to token-probability-based approaches, better aligns with downstream task performance.
By incorporating value models, rollouts, or backtracking, it also balances exploitation and exploration, enhancing sampling coverage.

\paragraph{Monte Carlo Tree Search}
Incontrast to BoN and traditional beam search, Monte Carlo Tree Search (MCTS) is a lookahead search algorithm used for making optimal decisions in large search spaces~\citep{browne2012survey}, which can choose actions based on their expected returns. 
Traditional MCTS use Monte Carlo estimations to estimate state values in a search tree.
As simulations increase, the search tree grows, values become more accurate, and the action-selection policy improves by prioritizing higher-value children.
MCTS is suitable for large-scale search spaces.
Compared to traditional algorithms, it efficiently explores complex search trees using estimated action values (expected rewards).
Moreover, MCTS balances exploration and exploitation, allowing the search to not only focus on the most promising areas but also explore unknown regions, thereby avoiding getting stuck in local optimal solutions.
As a result, it has shown strong performance on complex reasoning tasks with large search spaces, such as Go~\citep{AlphaGo}.

In MCTS, the nodes of the tree represent states, and the edges represent actions.
MCTS constructs a search tree through multiple MCTS simulations to estimate the value (expected reward) of candidate actions in the current state.
A typical MCTS simulation consists of four stages:

\begin{itemize}
    \item \textbf{Selection}: Starting from the root, the algorithm selects edges (actions) recursively based on the action values.
    To encourage exploration, MCTS adds an exploration term to the action values, which promotes actions that have been visited less frequently.
    This term is typically calculated using a variant of PUCT~\citep{variants_PUCT}.
    As a result, MCTS is able to explore more actions and make more accurate estimates of their values.
    After selecting an action, a new state is reached, which corresponds to a node in the tree.
    The selection process is repeated continuously until a leaf node is reached.
    \item \textbf{Expansion}: If the leaf node reached by the selection process is not a terminal node, such as not being the <eos> token in the text, MCTS will expand this node.
    It selects the possible actions for the leaf node's state and the resulting states after executing those actions as the child nodes of the current node, using the probabilities of these actions as the initial action values.
    Typically, MCTS expands the same number of actions each time.
    
    \item \textbf{Evaluation}: For the node reached by the selection process, MCTS evaluates the value of the node's state to refine the current action values.
    Traditional MCTS uses a more efficient policy for Monte Carlo estimation (e.g., using a smaller model), a process known as rollout, and the policy used for estimation is called the rollout policy~\citep{Sutton_book}.
    Another approach is to use an additional value model to predict the value of the current state without requiring extra sampling.
    In practice, rollout and the value model can be used together. 
    The advantage of this state value estimation is that MCTS, compared to traditional lookahead search algorithms, can more efficiently estimate the expected reward of the current state in the future without traversing the entire search tree, making it more suitable for problems with large search spaces.
    
    \item \textbf{Backpropagation}: After evaluating the value of the node's state, MCTS performs backpropagation to update the values of the actions along the path as well as the visit counts of the actions.
    Through backpropagation, MCTS refines its estimates of the action values, making the estimates more accurate.
\end{itemize}
The MCTS algorithm iteratively executes four stages until reaching a predefined simulation limit, selecting the most visited action from the root node.

MCTS algorithms based on LLMs mainly differ in action granularity and reward selection.
\citet{Alpha-zero-like} analyze the impact of token-level and step-level actions on the search space.
Step-level actions simplify the search process by reducing tree width but face challenges with large action spaces due to sentence diversity.
In contrast, token-level actions narrow the action space but increase tree depth.
Some approaches also adopt solution-level actions, where nodes represent entire solutions, and actions modify these solutions.

\begin{itemize}
    \item \textbf{Token-level actions: }
\citet{ppo_mcts} use a token-level action space, where the LLM predicts the next token as the MCTS action, with the PPO value model evaluating states.
\citet{KCTS} employ token-level MCTS to ground generation in reference knowledge, reducing hallucinations. 
\citet{MCTS_for_code_generation} use token-level MCTS to guide Transformers in generating better programs, evaluating node values via rollouts.
Accurately assessing token values is critical, but sparse reward signals pose challenges in deriving token-level value or reward models.
Additionally, the deep search tree leads to efficiency issues, making the design of an efficient inference framework essential for token-level MCTS.

\item \textbf{Step-level actions:}
For multi-step reasoning tasks, step-level actions are a natural choice.
\citet{RAP} propose RAP, an MCTS-based planning algorithm, which defines actions based on task-specific steps, such as posing sub-questions in math problems or moving blocks in Blocksworld~\citep{Blocksworld}.
RAP uses an LLM to predict future states instead of observing state transitions.
\citet{AlphaMath} treat partial solutions as states and reasoning steps as actions for mathematical problems.
\citet{AgentQ, Treesearch_unifies_reasoning_acting_planning} use browser interaction commands as actions in WebShop, a simulated e-commerce platform.
Unlike task-specific steps, \citet{rStar} design five human-like reasoning actions for MCTS.
To achieve o1-style reasoning, we may need to manually design actions like reflection and error correction, which should be incorporated during policy initialization.

\item \textbf{Solution-level actions:}
Solution-level MCTS treats the complete solution as a node state and modifications as actions.
\citet{MCTS-r} propose MCTSr, where nodes represent different answer versions, actions involve self-refinement, and rewards come from self-evaluation.
Starting with a naive model-generated answer, MCTSr combines self-refinement with MCTS to achieve GPT-4-level performance on math problems.
Building on MCTSr, \citet{llama-berry} use a pairwise preference reward model to evaluate whether a modified answer improves, estimating the new node's value.
\citet{Mirror} use self-improvement as actions and design a consistency-based method to ensure factual correctness.

\end{itemize}

\paragraph{Other Tree Search Algorithms.}
In addition to MCTS and Beam Search, traditional graph search algorithms like DFS, BFS, and $A^*$ are also used.
The Tree-of-Thought (ToT) framework~\citep{yao2023tree} models reasoning as a tree, where nodes represent reasoning steps and branches represent continuations.
By expanding and evaluating multiple reasoning solutions, ToT explores a broader solution space, introducing mechanisms like search, reflection, and backtracking to reconsider steps and explore alternatives.
ToT employs DFS and BFS for its search process.
\citet{Tree_search_for_language_model_agents} propose an $A^*$-inspired best-first search, using a multimodal LLM to evaluate nodes and a heuristic to select or backtrack.
\citet{Beyond_A*} use $A^*$ to collect execution traces, training an encoder-decoder model to imitate these traces.

\subsubsection{Sequential Revisions}
\label{subsubsection:sequential_revisions}
Compared to tree search, sequential revisions primarily conducts the search by iteratively refining the previous answer.
The key feature of sequential revisions is that it generates improved answers based on reflections on previous answers or changes in the environment.
Sequential revisions requires the model to have basic abilities for self-reflection and error correction, which can be introduced during the poicy initialization through SFT or Prompting, as discussed in Section \ref{policy_initilization}.

Sequential revisions can directly rely on the internal guidance such as self-evaluation.
\citet{Self_refine} propose to SELF-REFINE.
It first generates an initial answer using a LLM, and then utilize the same LLM to iteratively provide feedback for its output and refine its output based on the self feedback.
The iteration process stops when the maximum number of iterations is reached or when the model determines that it can stop.
\citet{Scaling_test_time_compute} also uses a model with revision capabilities to perform revisions during inference, demonstrating that the accuracy of the answer improves as the number of revisions increases.
Other work rely on observation or feedback from external environment.
Both \citet{ReACT} and \citet{Reflexion} are based on the model reflecting on or improving the next action based on external feedback observed after taking an action.
\citet{self_debug} and \citet{Critic} use feedback based on the execution of generated code to allow the model to modify the generated code or debug.

There is still debate over whether Sequential Revisions is truly effective.
For example, \citet{LLM_cannot_self_correct} argue that large models cannot self-correct properly without external feedback.
However, the opposing viewpoint suggests that due to the existence of the DG-gap~\citep{janleike2022why_excited_about_AI_assisted_human_feedback}, large models may have a stronger ability to discern and improve the answers they have already generated, allowing for further refinement.
\citet{Tree_search_vs_revisions} conduct empirical studies across a range of tasks and found that Sequential Revisions only achieves performance improvements over simpler methods like BoN when the accuracy of the discriminator (guidance) $\geq$ 90\%.

\begin{table*}[ht]
\centering
\footnotesize
\begin{adjustbox}{width=\textwidth,center}
\begin{tabular}{p{5cm}p{2.4cm}p{2.4cm}p{2cm}p{2.6cm}}
\toprule
Paper & \multicolumn{2}{c}{Search Guidance} & \multicolumn{2}{c}{Search Strategies} \\ 
\cmidrule(lr){2-3} \cmidrule(lr){4-5}
      & Internal Guidance & External Guidance & Tree Search & Sequential Revisions \\
\midrule
Math Verifier~\citep{OpenAIMathVerifierORM} & \xmark & Env Feedback & Best-of-N  & \xmark \\
\rowcolor{Gray}
Self-consistency~\citep{SC} & Model Uncertainty & \xmark & Best-of-N & \xmark \\
Speculative Rejection~\citep{Speculative_bon} & \xmark & Env Feedback & Best-of-N & \xmark \\ 
\rowcolor{Gray}
BoN~\citep{BoNBoN},vBoN~\citep{Variational_bon},BOND~\citep{BOND} & \xmark & Env Feedback & Best-of-N\(^\dagger\) & \xmark \\ 
TreeBoN~\citep{TreeBoN} & \xmark & Env Feedback & Beam Search & \xmark \\ 
\rowcolor{Gray}
OVM~\citep{OVM} & Value Function & Value Function & Beam Search & \xmark \\
\citet{Self_evaluation_guided_beam_search} & Self-evaluation & \xmark & Beam Search & \xmark \\
\rowcolor{Gray}
\citet{Reward_step_by_step} & \xmark & Env Feedback & Beam Search\(^\ddagger\) & \xmark \\
\citet{Scaling_test_time_compute} & \xmark & Env Feedback & Beam Search & \xmark \\
\rowcolor{Gray}
TS-LLM~\citep{Alpha-zero-like} & \xmark & Env Feedback & MCTS & \xmark \\
PPO-MCTS~\citep{ppo_mcts} & Value Function & Value Function & MCTS & \xmark \\
\rowcolor{Gray}
KCTS~\citep{KCTS} & \xmark & Env Feedback & MCTS & \xmark \\
\citet{MCTS_for_code_generation} & \xmark & Env Feedback & MCTS & \xmark \\
\rowcolor{Gray}
RAP~\citep{RAP} & Self-evaluation & Heuristic Rules & MCTS & \xmark \\
AlphaMath~\citep{AlphaMath} & Value Function & Value Function & MCTS,\newline Beam Search & \xmark \\
\rowcolor{Gray}
AgentQ~\citep{AgentQ} & Self-evaluation & \xmark & MCTS & \xmark \\
LATS~\citep{Treesearch_unifies_reasoning_acting_planning} & Self-evaluation, \newline Model Uncertainty & \xmark & MCTS & \xmark \\
\rowcolor{Gray}
rStar~\citep{rStar} & Model Uncertainty & \xmark & MCTS & \xmark \\
MCTSr~\citep{MCTS-r} & Self-evaluation & \xmark & MCTS & \cmark \\
\rowcolor{Gray}
LLaMA-Berry~\citep{llama-berry} & \xmark & Env Feedback & MCTS & \xmark \\
\rowcolor{Gray}
Mirror~\citep{Mirror} & \xmark & Heuristic Rules & MCTS & \cmark \\
ToT~\citep{yao2023tree} & \xmark & Env Feedback,\newline Heuristic Rules & DFS,BFS & \xmark \\
\rowcolor{Gray}
\citet{Tree_search_for_language_model_agents} & \xmark & Env Feedback & $A^*$ & \xmark \\
Beyond $A^*$~\citep{Beyond_A*}\(^\ddagger\) & \xmark & Heuristic Rules & $A^*$ & \xmark \\
\rowcolor{Gray}
Self-refine~\citep{Self_refine} & Self-evaluation & \xmark & \xmark & \cmark \\
\citet{Scaling_test_time_compute} & Self-evaluation & \xmark & \xmark & \cmark \\
\rowcolor{Gray}
ReACT~\citep{ReACT} & \xmark & Env Feedback & \xmark & \cmark \\
Reflexion~\citep{Reflexion} & \xmark & Env Feedback & \xmark & \cmark \\
\rowcolor{Gray}
Self-debug~\citep{self_debug} & \xmark & Env Feedback & \xmark & \cmark \\
Critic~\citep{Critic} & \xmark & Env Feedback & \xmark & \cmark \\
\bottomrule
\end{tabular}
\end{adjustbox}
\caption{Survey of existing search methods, including their search guidance and search strategies. \(^\dagger\): These work finetune models to simulate BoN distributions. \(^\ddagger\): This work use greedy search, which is a special case of beam search. Env Feedback means Environmental Feedback.
Value function is a combination of internal and external guidance, thus it appears in both the internal and external guidance columns.}
\label{tab:search_survey}
\end{table*}
As a summary, we list the research work in Table \ref{tab:search_survey}, along with the specific categories of guiding signals and search strategies used, for the reader's reference.

\subsubsection{Comparison of Tree Search and Sequential Revisions}
As we mentioned in Section \ref{sec:tree_search}, tree search typically samples multiple candidate solutions simultaneously, which allows some tree search algorithms, such as BoN, to be accelerated using parallel strategies.
At the same time, the candidate solutions in tree search are relatively independent of each other, and the search process can encourage exploration during sampling, such as in MCTS.
This broadens the coverage of candidate solutions.
On the other hand, sequential revisions generally continue to search around a previously sampled solution, allowing for, incremental improvements, which may lead to better solutions in some cases.
For example, \citet{Scaling_test_time_compute} show that sequential revisions with internal guidance outperform majority voting (a tree search strategy with internal guidance).
However, since sequential revisions rely on previous attempts, the computational cost increases with the number of revisions made.

\subsubsection{Combination of Tree Search and Sequential Revisions}

Tree search and sequential revisions can be used together.
In tree search, using solution-level search nodes can be seen as a combination of tree search and sequential revisions~\citep{MCTS-r, llama-berry}.
Additionally, \citet{Scaling_test_time_compute} combines BoN with sequential revisions by first randomly sampling N candidate solutions, then applying sequential revisions to these N solutions, and finally using a verifier to select the best solution from all of them.
This combined approach achieves performance that surpasses BoN.
Such results may demonstrate the potential of combining these two search strategies.

\subsection{Speculation About the Search of o1.}\label{sec:diff-train-and-search}
In this section, we discuss the guiding signals and search strategies used during both the training-time search and inference-time search in o1.

\paragraph{Train-time search.}
During training, o1 is more likely to employ tree search techniques, such as BoN or tree search algorithms, and primarily relies on external guidance.
This is because the model needs to gradually enhance its reasoning capabilities during training, and tree can sample a large number of candidate solutions in parallel, efficiently providing the model with abundant high-quality training data.
Furthermore, since real-time interaction is not required during training, various external environments can be accessed to validate the sampled solutions, such as executing code or verifying the accuracy of mathematical computations.
This external guidance helps more accurately direct the model's search process.

\paragraph{Test-time search.}
For test-time search, o1 is more likely to use sequential revisions, combining internal guidance to continuously refine and correct its search through reflection.
Through the example in the o1 blog~\citep{o1_blog}, we can observe that o1's reasoning style is closer to sequential revisions. 
Besides, using Tree Search for long reasoning processes can result in significant overhead.
And during inference, it is difficult to rely on real-world environments for guidance, and~\citep{RewardOverOptim, inverse_inference_time_scaling} point out that performing extensive searches based on proxy feedback, like reward models, can lead to overoptimization problem.
While increasing the computations during inference, it can actually degrade performance.
However, observations from o1's blog suggest that the model's performance continues to improve as the computational effort during inference increases.
This leads us to believe that o1 primarily uses internal guidance during the inference stage.
The computations during inference is mainly reflected in the length of the inference chain.

\subsection{Scaling Law of Search}
Prior to OpenAI's release of o1, several studies explored inference-time scaling laws.
\citet{Large_Language_Monkeys} studied BoN, dividing BoN into two stages: sampling solutions and selecting the best one via a verifier.
They found that increasing the number of samples improved "coverage" (pass@1 accuracy) following a power law.
Small models could achieve near 100\% pass@1 accuracy on MATH with sufficient scaling.
However, they noted a gap between best-of-n accuracy and pass@1, attributing it to the verifier's limitations in identifying correct solutions.

\citet{Scaling_test_time_compute} analyzed the scaling of tree search and sequential revisions, showing both approaches scaled well with more computational resources.
They found that, for a given computational budget, increasing model size was more effective for complex tasks, while sampling more tokens sufficed for simpler tasks.
\citet{VerifySbyS} examined reward signals in scaling and observed that outcome-reward-based best-of-n search plateaued as samples increased, while PRM-based best-of-n search avoided this issue.

While these studies showed performance gains with increased search computation, \citet{RewardOverOptim} found the \textbf{inverse scaling law} where scaling best-of-n search could degrade performance due to distribution shift.
The reward model, trained on original policy data, struggles to generalize to new policy.
\citet{inverse_inference_time_scaling} noted similar issues and suggested that other search algorithms, like MCTS, may also face inverse scaling challenges.

\subsection{Challenges of Search for Reproducing o1}
\paragraph{How to overcome Inverse Scaling}
One approach is to reduce test-time search, as the inverse scaling phenomenon occurs primarily during large-scale search.
However, this limits the scale of search. Alternatively, improving the reward model's generalization to handle unseen states is another solution.
Inspired by LLM development, this can be achieved by increasing the model's size and training data, as validated in~\citet{RewardOverOptim}.

\paragraph{How to avoid over-thinking on simple tasks?}
Not all problems require complex reasoning or search. For straightforward queries like ``1+1=?'', engaging in elaborate analysis wastes computational resources and potentially introduces errors. 
Forcing reasoning on such problems wastes resources and causes delays.
To address this, we could constrain the length of the chain of thought (CoT) using reward shaping with a length penalty.
This reshaped reward balances minimizing superfluous search and solving problems effectively.

\paragraph{How to trade off tree search and sequential revisions?}
Search scales across two dimensions: tree search and sequential revision.
Combining both improves performance~\citep{marco-o1}, but with a fixed computational budget, the optimal allocation of resources remains unclear.
This challenge is similar to balancing model size and data size under a fixed budget.
Empirical scaling laws can provide guidance for resource allocation.

\paragraph{How to improve efficiency of Search?}
A key challenge in scaling search is efficiency, as LLMs' auto-regressive generation is limited by memory read-write speeds, restricting GPU utilization~\citep{flash-attention}.
Additionally, some tree search algorithms, like MCTS, lack inherent parallelism.
Improving efficiency requires both engineering and algorithmic solutions.
For example, engineering optimizations involves implementing KV-cache sharing, while algorithmic improvements can perform KV-cache compression \citep{streamingllm,mem-step-by-step} and speculative sampling \citep{sq-sampling}.

\section{Learning}
\label{learning}
We have introduced policy initialization, which learns from human-expert data \footnote{policy initialization can also learn from a stronger model, which we ignores here for simplicity.},
why do we still need reinforcement learning? What makes reinforcement learning essential is that the training data of reinforcement learning is unlimited which comes from the interaction with an environment. In contrast, the human-expert data is limited and expensive. Furthermore, reinforcement learning has the potential to achieve \textbf{superhuman} performance, since it learns from trial and error instead of human-expert data. While human-expert data captures human behavior and knowledge, reinforcement learning can lead to the discovery of strategies that humans may not be capable of. AlphaGo \citep{AlphaGo} which utilized reinforcement learning, was able to defeat world-class human players in the game of Go by discovering novel strategies that were previously unknown to experts, for example the famout ``37 move'' of AlphaGo \citep{AlphaGo}

The reinforcement learning typically samples the trajectory using a policy and improves the policy based on the received rewards. In the context of o1, we hypothesize that the reinforcement learning process generates trajectories through a search algorithm rather than relying solely on sampling. One advantage of search methods is their ability to explore superior states or solutions compared to random sampling. For example, beam search prioritizes actions with the highest expected action values. Thus, search techniques can provide higher-quality training data than simple sampling. Under this assumption, the reinforcement learning of o1 could involve an iterative process of search and learning. In each iteration, the learning phase utilizes the search-generated outputs as training data to enhance the policy, while the improved policy is then applied to the search process in the next iteration. A prominent example of this search-and-learn iteration is AlphaGo Zero \citep{AlphaZero}, which uses trajectory data obtained via Monte Carlo Tree Search (MCTS) for policy learning.

\begin{figure}
    \centering
    \includegraphics[width=\linewidth]{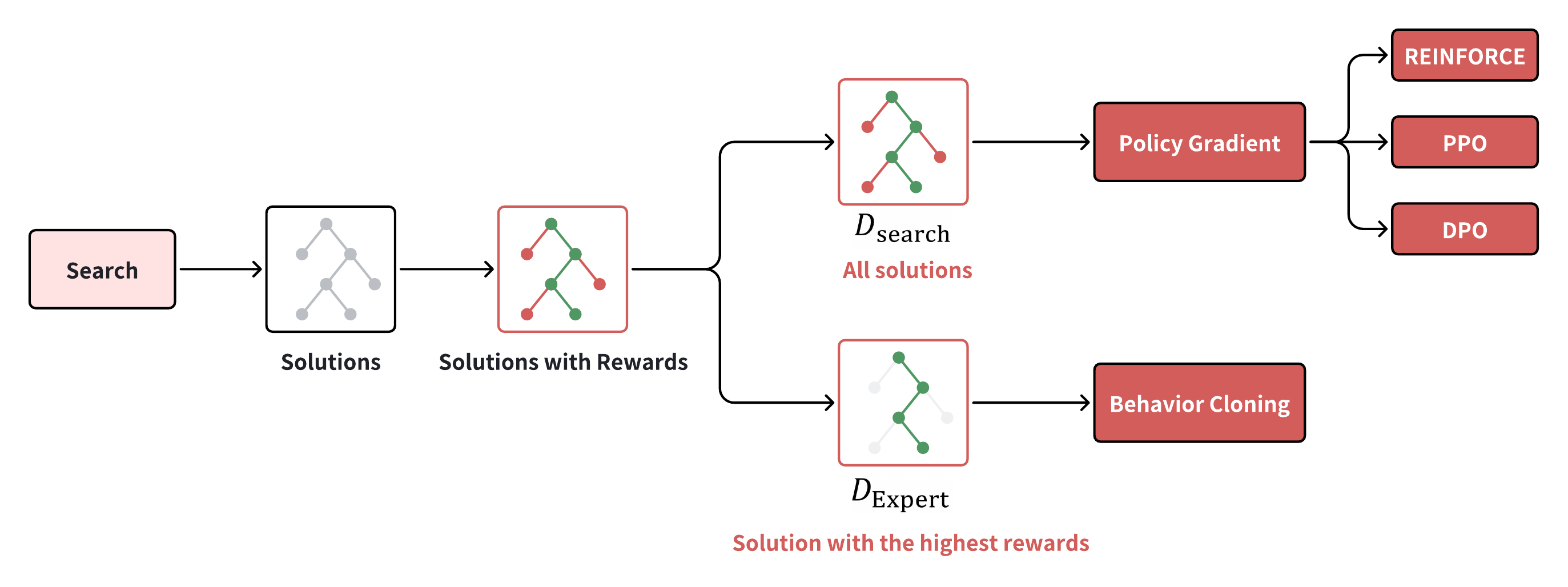}
    \caption{The difference of 4 learning methods, in terms of the data they use. Policy gradient use $D_{\text{search}}$ which contains all searched solutions, while behavior cloning only utilizes $D_{\text{expert}}$, the solutions with the highest rewards. We distinguish the solution with the highest reward from the other solutions by using different colors: green for the highest reward and red for the others.}
    \vspace{-1em}
    \label{fig:search_data}
\end{figure}

The train-time search is different from the test-time search. The test-time search outputs a solution with the maximum rewards or confidence in all candidate solutions. But at the training, all candidate solutions generated from search may be utilized by learning. We denote the set of state-action pairs output from search as $D_{\text{search}}$, and the set of state-action pairs in the optimal solutions from search as $D_{\text{expert}}$. Therefore, $D_{\text{expert}}$ is a subset of $D_{\text{search}}$. We visualize the difference between $D_{\text{search}}$ and $D_{\text{expert}}$ in Figure \ref{fig:search_data}. 

\subsection{Learning Methods}
Given \( D_{\text{search}} \), the policy can be improved using either policy gradient methods~\citep{reinforce} or behavior cloning. This section provides a detailed overview of several key policy gradient methods. Among these, Proximal Policy Optimization (PPO) \citep{reinforce} and Direct Policy Optimization (DPO) \citep{dpo} are the most widely used reinforcement learning techniques for LLMs. It is also common to preform behavior cloning or supervised learning on the searched data \citep{Star,llama2}. The behavior cloning here differs from that in policy initialization discussed in Section \ref{policy_initilization} in that it learns the behavior derived from the search process, rather than from human expert.

\subsubsection{Policy Gradient}
The policy gradient parameterizes the policy with $\theta$, which can be defined as $\pi_{\theta}(a|s)$. $\theta$ is updated with the following gradient:

\begin{equation}
   \nabla_\theta J(\theta) = \mathbb{E}_{\pi_{\theta}} \left[  Q(s_t,a_t) \nabla_\theta \log \pi_\theta(a_t|s_t) \right]
    \label{eq:j_theta}
\end{equation}
Q-value $Q(s_t, a_t)$ is often approximated with Monte Carlo sampling \citep{reinforce} or Temporal-Difference estimation \citep{actor-critic}. 

\paragraph{REINFORCE} REINFORCE algorithm~\citep{reinforce} uses Monte Carlo to approximate action value:
\begin{equation}
    \nabla_\theta J(\theta) = \frac{1}{|D_{\text{search}}|}\sum_{(s_t,a_t) \in D_{\text{search}}} \left[  G_t \nabla_\theta \log \pi_\theta(a_t|s_t) \right],
\end{equation}
where the return $G_t$ is the discounted cumulated rewards, $D_{\text{search}}$  denotes all state-action pairs in the searched data. 

\paragraph{Actor-Critic} The return $G_t$ may bring high variance for the gradient estimation. Therefore, Actor Critic~\citep{actor-critic} replace $G_t$ with the advantage function or Temporal-difference error, which defined $A(s_t,a_t)=R_{t+1} + \gamma V_{\pi_{\theta}}(s_{t+1})-V(s_{t})$.

The advantage function in Actor-Critic algorithm involves the estimation of value function $V(S)$, therefore it learns a value model to approximate the value function. Another benefit of using Temporal-difference error $A(s_t,a_t)$ instead of $G_t$ is that the estimation of $G_t$ may require more computation for sampling. For example, some trajectories in beam-search are pruned before reaching the terminal state. We need to make additional sampling to estimate $G_t$ for these intermediate states.

\paragraph{PPO} In policy gradient methods, the policy is updated using gradient ascent. However, when the policy update is too large, it can cause drastic changes in the policy, leading to poor performance or even divergence. TRPO \citep{trpo} and PPO \citep{ppo} address this problem with KL divergence constraint and clipping, which ensuring that policy updates are not too large. 

Another benefit of TRPO and PPO is that it improves data utilization. It is more data-efficient to update the parameters $\theta$ multiple times on the same example, like the multi-epoch training in machine learning. Repeatably learning on the same data is called Replaying buffer in reinforcement learning. Replaying buffer improves the data utilization, but making the learning off-policy, where the behavior policy ($\pi_{\text{search}}$) that generates $D_{\text{search}}$ is not the same as the target policy ($\pi_\theta$) that is to be optimized. TRPO~\citep{trpo} and PPO
~\citep{ppo} alleviates this problem with KL divergence constraint and clipping.

PPO is a variant of Actor-Critic algorithm, which has been widely used for reinforcement learning for large language model, especially the RLHF~\citep{InstructGPT,rlhf-anthropic}. But the training of PPO typically involves 4 models for function approximation of $\pi_{\theta}$, $\pi_{\text{search}}$, reward, value, which is expensive. GRPO~\citep{shao2024deepseekmathpushinglimitsmathematical} alleviate this issue by estimating value function with Monte Carlo estimation instead of function approximation. Remax~\citep{remax} is also proposed to simplify the training of PPO, which use the return of greedy decoding as the baseline of reduce the variance of reinforce algorithm.  

It is not trivial to reimplement the performance of PPO,~\citet{ppo-implementation} highlight 13 core implementation details of PPO, for example, using two distinct networks to predict policy and value respectively is better than sharing their parameters.~\citet{rlhf-secret-1,rlhf-secret-2} also empirically validate the influence of some implementation details on the performance of RLHF, for example,  they  found that it is helpful to normalize and clip reward.

\paragraph{DPO}
DPO eliminate the cost of modeling value model and reference policy, meanwhile having low variance in reward/value estimation. The idea of DPO is to reparameterize reward function with the optimal policy $\pi^*$, and transform the optimization of Bradley-Terry reward modeling to the optimization of the policy. The loss function of DPO is:
\begin{equation}
\mathcal{L}_{\text{DPO}}(\pi_{\theta}; \pi_{\text{ref}}) = -\mathbb{E}_{(x,y_w,y_l) \sim \mathcal{D_{\text{search}}}} \left[ \log \sigma \left( \beta \log \frac{\pi_\theta(y_w | x)}{\pi_{\text{ref}}(y_w | x)} - \beta \log \frac{\pi_\theta(y_l | x)}{\pi_{\text{ref}}(y_l | x)} \right) \right]
\label{eq: dpo}
\end{equation}

where $x$ denotes the question, $y_w$ denotes the positive solution and $y_l$ denotes the negative solution. The parametrization of DPO can derive a kind of reward shaping:
\begin{equation}
    f(r, \pi_{\text{ref}}, \beta)(x, y) = r(x, y) - \beta \log \sum_{y} \pi_{\text{ref}}(y | x) \exp\left(-\frac{r(x, y)}{\beta}\right)
\end{equation}
Here, the term \(\beta \log \sum_{y} \pi_{\text{ref}}(y | x) \exp\left(-\frac{r(x, y)}{\beta}\right)\) serves as the baseline for \(r(x, y)\), which explains why DPO does not encounter the high variance issue associated with the REINFORCE algorithm. Estimating this baseline is computationally expensive due to the large action space, but DPO eliminates this estimation by leveraging the Bradley-Terry model.

DPO has been widely used as the alternative for PPO in RLHF~\citep{llama3}. Even though the search does not provide the preference data directly, we can still construct preference pairs from $D_{\text{search}}$. LLama-3~\citep{llama3} uses the positive and negative examples from rejected sampling as the preference data for DPO. MCTS-DPO~\citep{mcts-dpo} and SVPO~\citep{svpo} collect a number of step-level preference pairs from state-action pairs from MCTS, and use the preference data for DPO. 

\subsubsection{Behavior Cloning}
Behavior cloning, or supervised learning, employs the expert policy as the target and seeks to reduce the discrepancy between the policy \(\pi_{\theta}\) and the expert policy \(\pi_{\text{expert}}\). While \(\pi_{\text{expert}}\) may not represent a true expert policy, it is generally more effective than \(\pi_{\theta}\). Furthermore, \(\pi_{\text{expert}}\) improves with increased computational effort, indicating that, with sufficient resources, \(\pi_{\text{expert}}\) could approach the performance of a true expert policy.
For LLM, the loss function of behavior cloning is cross-entropy loss.
$$
\text{min}_{\theta} -\mathbb{E}_{(s, a) \sim \pi_{\text{expert}}} \left[ \log \pi_{\theta}(a|s) \right]
$$

 $D_{\text{search}}$ contains all state-action pairs from search, including the negative solution, therefore it can not be taken as the expert data. In contrast, the state-action pairs in $D_{\text{expert}}$ receive the highest rewards compared with other pairs. Therefore, $D_{\text{expert}}$ can be taken as the expert data. The expert data from search may not really have a high quality, but it is better than the data sampled from the current policy. And as we scale the compuation of search, the expert data generated from search becomes more optimal. For example, \citet{RewardOverOptim} found scaling best-of-n search enlarges the distance between policy of the best-of-n solution and the current policy.

\begin{equation}
    \text{min}_{\theta} -\frac{1}{|D_{\text{expert}}|}\sum_{(s, a) \in D_{\text{expert}}} \left[ \log \pi_{\theta}(a|s) \right],
\end{equation}

The approach of iterating search and behavior cloning is called Expert Iteration in ~\citet{expert-iteration}. AlphaGo Zero~\citep{AlphaZero} is a well-known example of expert iteration, which uses the policy of MCTS for behavior cloning. 

Although Expert iteration uses behavior cloning which can be seen as the supervised learning, it is different from policy initialization introduced in Section \ref{policy_initilization}, where the expert data comes from human expert or a stronger model. The expert data of expert initialization comes from the interaction with the environment, i.e., search.

A well-known example of expert iteration for large language model is STaR~\citep{Star} that uses reject sampling as the search algorithm to filter the examples generated by LLM with incorrect answers and finetune LLM on the remaining examples. LLama-2~\citep{llama2} utilizes the method of STaR for RLHF and found it achieve the comparable performance with PPO. V-Star~\citep{v-star} follows Star to use reject sampling to filter high-quality examples, but they train a verifier using DPO to provide reward for reject sampling rather than relying on the ground-truth answer. Their verifier also works at the test-time, by applying on best-of-n search.~\citet{Alpha-zero-like,alphazero-like-learning2,AlphaMath,alphazero-like-learning4} follows AlphaGo Zero to use MCTS as the search algorithm and finetune LLM on the solutions with the highest or the top-k highest reward in the tree. The main difference of their work between STaR is that they used different search algorithm.

\subsubsection{Speculation about the learning of o1}
\label{sec:speculate-learning}
\begin{table}[t]
    \centering
    \resizebox{\textwidth}{!}{
    \begin{tabular}{c|c|ccc|cc}
        \toprule
        \multirow{2}{*}{\parbox{2.5cm}{\centering Learning Method}} & 
        \multirow{2}{*}{\parbox{2.5cm}{\centering Gradient Variance}} & 
        \multicolumn{3}{c|}{Memory Cost} & 
        \multicolumn{2}{c}{Data Utilization} \\
         & & Reward Model & Value Model & Reference Policy & Replay Buffer & Negative solutions \\
        \midrule
        REINFORCE & {High} & \checkmark & \blackxmark & \blackxmark  & \blackxmark  & \checkmark \\
        PPO & {Low} & \checkmark & \checkmark & \checkmark  & \checkmark & \checkmark \\
        DPO & {Low} & \blackxmark & \blackxmark & \checkmark  & \checkmark & \checkmark \\
        \hline
        Behavior Cloning & {Low} & \blackxmark & \blackxmark & \blackxmark & \checkmark & \blackxmark \\
        \bottomrule
    \end{tabular}
    }
    \caption{A comparison of 4 learning methods—REINFORCE, PPO, DPO, and behavior cloning—can be made based on three factors: gradient variance, memory cost, and data utilization. For memory cost, \checkmark means requiring reward/value/reference model, suggesting high memory cost. As for data utilization, \checkmark means using replaying-buffer/negative-solutions for learning, which indicates high data utilization.}

    \label{tab:learning-compare}
\end{table}

We make some comparison on the PPO, DPO and behavior cloning to see which one is more likely to be used for o1. The differences between the three methods in terms of memory cost and data utilization are summarized in Table \ref{tab:learning-compare}. 

\paragraph{Memory Cost} PPO requires storing reward function, value function and reference policy in memory, which is expensive. While DPO eliminates the reward model and value model, which is simpler and more memory-efficient than PPO. But DPO is based on Bradley-Terry model and requires the preference data. For some environment, where the ground truth is available, it would be better to learn a policy with PPO using the reward signal, instead of learning from preference data. Behavior cloning even does not need reference policy, thus is the most memory-effient among three learning methods.

\paragraph{Data Utilization} The difference among the 3 learning methods is not only on the learning algorithm, but also on the training data. PPO and DPO use all state-action pairs from search ($D_{\text{search}}$), even those with negative rewards. While behavior cloning adopts the subset of state-action pairs from search with high rewards ($D_{\text{expert}}$). Therefore, PPO and DPO have better data utilization than behavior cloning, since the negative action or solution can also provide useful signal for improving policy. 

~\citet{compare-expert-iter-ppo} empirically compare the performance between PPO \citep{ppo} and behavior cloning on GSM8K and MATH, and they find the behavior cloning (expert iteration) consistently outperforms policy gradient. It requires more empirical comparison on these methods.

It is likely that the learning of o1 results from the combination of multiple learning methods. In this framework, we hypothesize that the learning process for o1 begins with a warm-up phase using behavior cloning, followed by a transition to PPO or DPO once the improvements from behavior cloning plateau. This approach is grounded in the idea that behavior cloning is more efficient than PPO or DPO, thereby accelerating the warm-up phase. However, behavior cloning is limited in that it only learns from the highest-reward solutions and disregards negative solutions. Consequently, further optimization may require the use of PPO or DPO, which provide better data utilization. This pipeline is consistent with the post-training strategies employed in LLama2 \citep{llama2} and LLama3 \citep{llama3}.

\subsection{Scaling Law of Reinforcement Learning}
The relationship between the loss, computational cost, model parameters, and data size during the pre-training phase of large models follows a power-law scaling law. But does reinforcement learning (RL) in large language models also exhibit a scaling law? The blog of OpenAI \citep{o1_blog} shows that there is a log-linear scaling law between reasoning performance and train-time compute. However, aside from this, there has been little research on the scaling laws related to reinforcement learning.

OpenAI has also investigated the scaling laws in reinforcement learning \citep{rl-scale-law}, their experiments focused on traditional reinforcement learning tasks, such as Dota 2, and did not involve LLMs. They found that the performance of reinforcement learning is improved with the increase of the number of model parameters and interactions with environment, which follows power law. They were able to derive the optimal number of model parameters and interactions with environment under a fixed computational budget. 

~\citet{imitation-learn-scale-law} studied the scaling laws of imitation learning in traditional reinforcement learning tasks, such as Atari. They also observed a power-law relationship between the imitation-learning loss and scale factors, including model size, data size, and computation budget. However, the data used for imitation learning came from expert annotations, rather than being generated through interactions with environment.

Some studies on reinforcement learning with large models have experimentally explored the relationship between LLM performance and the number of iterations~\citep{Star,mcts-dpo}. However, these studies only conducted small-scale validations, showing that scaling the number of iterations is effective, but they did not establish scaling laws that could be used to predict results for larger-scale experiments. In order to conduct large-scale reinforcement learning like o1, it is crucial to investigate the scaling laws of reinforcement learning on LLMs.

\subsection{Challenges of Learning for Reproducing o1}

\paragraph{How to improve Training Efficiency?}
The primary bottleneck in training efficiency arises from the train-time search process, as the time required for LLM generation on the same batch exceeds that of training, with search being particularly slow. For instance, in the open-source project MCTS-DPO \citep{mcts-dpo}, the majority of the training time is consumed by the MCTS search, leading to training times of up to one week on an A800 GPU using the MATH dataset. There are two potential strategies for accelerating training: 1. Improving the search algorithm and implementation, which has been discussed in the search section and will not be reiterated here. 2. Extending learning beyond the data generated by online search to include data from previous search iterations. While reusing data from previous iterations may introduce issues related to off-policy learning, it increases data utilization and consequently reduces the search scale.

\paragraph{How to learn a Strong Question Generator.}
We have introduced how the solutions generated by search can be used for learning. But the data for learning includes not only the solution but also question. As the policy of LLM improves, the challenging questions become simple. Therefore, it may be important to update the question or initial states. For example, the new questions can be more challenging or explore the new domains. Wizzrd-LM \citet{WizardLM} proposes to prompt a strong LLM like chatgpt to generate a more challenging question or another question that is related to the given question. But the generation of new question is independent from the performance of the policy to be optimized. The problem of updating questions is also related to Curriculum Learning \citep{cl-survey}, especially automatic curriculum learning \citep{automatic-cl}. Generating questions conditioned on the policy for LLM is challenging, since the generated questions may not be suitable or even unsolvable. Learning on these questions are not helpful for improving policy of LLM.

\paragraph{How to Narrow the Distribution Shift in off-policy learning?}
The solutions generated by search is typically better than the data sampled from the current policy. The result is that the data produced by search can be considered as originating from a better policy. Using search-generated data for policy gradient training constitutes off-policy learning. To mitigate the distribution shift of policy in off-policy learning, one straightforward approach is to limit the scale of the search. The smaller the search scale, the less pronounced the distribution-shift issue becomes. Alternatively, we can implement search with sampling from the current policy, then there is no distribution shift problem.

An alternative approach is to incorporate off-policy learning methods. While techniques such as importance sampling and KL divergence constraints, employed in TRPO~\citep{trpo} and PPO~\citep{ppo}, are effective. However, they require knowledge of the policy probability associated with the search data, denoted as $\pi_{\text{search}}$, which is unavailable.

Another approach to make use of behavior cloning to turn the off-policy learning to on-policy learning. We can first apply behavior cloning on searched data \(D_{\text{search}}\) to narrow the gap between$\pi_{\theta}$ and $\pi_{\text{search}}$. After completing the behavior cloning, policy gradient training can then be conducted on the searched data. The pipeline of policy gradient following behavior cloning occurs at each iteration, which is different the speculation introduced in \ref{sec:speculate-learning} , where we perform behavior cloning at early iterations and PPO at the remaining iterations. Furthermore, this approach can be supplement for our speculation: We perform behavior cloning at the warm-up stage, and then combine behavior cloning and PPO at each iteration.

\section{Open-source o1 Project}
\label{open-source-o1}
\begin{table}
    \centering
    \resizebox{\textwidth}{!}{
    \begin{tabular}{ccccccl}
        \toprule
        & & & \multicolumn{2}{c}{Reinforcement Learning} &  &\\
        \cmidrule(r){4-5}
        Project & Initialization & Reward Design & Train-time Search & Learning & Test-Time Search  &Resource\\
        \midrule
         g1& Prompt& -& -& -&Sampling & \href{https://github.com/bklieger-groq/g1}{Prompt}\\\rowcolor{Gray}
         Thinking Claude& Prompt& -& -& -&Sampling & \href{https://github.com/richards199999/Thinking-Claude}{Prompt} \\
         Open o1& SFT& -& -& -&Sampling & \href{https://github.com/Open-Source-o1/Open-o1}{Data} \\\rowcolor{Gray}
        o1-journey (part 1) & SFT& PRM & Beam-Search & Behavior Cloning & Sampling & -\\
        o1-journey (part 2) & SFT& - & - & - & Sampling & -\\\rowcolor{Gray}
        Open-Reasoner & -& PRM & Sampling& PPO & MCTS  & \href{https://github.com/openreasoner/openr}{Code} \\
        Slow Thinking with LLMs 1& SFT& ORM & Sampling& DPO & MCTS & - \\\rowcolor{Gray}
        Slow Thinking with LLMs 2& SFT& ORM & Sampling& DPO/SFT& smapling& - \\
        Marco-o1 & SFT& ORM & MCTS & Behavior Cloning & MCTS & \href{https://huggingface.co/AIDC-AI/Marco-o1}{Model}\\\rowcolor{Gray}
        o1-coder& SFT& PRM& MCTS& PPO/DPO&MCTS & - \\
        \bottomrule
    \end{tabular}
    }
    \caption{Comparison of different open-source o1 projects.}
    \label{tab:open-source-o1}
\end{table}

Although o1 has not published a technical report, the academic community has made available several open-source implementations of o1. All of these implementations can be viewed as components or special cases of the o1 framework introduced in this paper. In Table \ref{tab:open-source-o1}, we summarize the methods adopted by these open-source projects in the areas of  policy initialization, reward design, search, and learning.

There are also some o1-like models from industry, for example, k0-math, skywork-o1 \citep{skywork-o1}, Deepseek-R1 \citep{shao2024deepseekmathpushinglimitsmathematical}, QwQ \citep{qwq} and InternThinker \citep{cai2024internlm2}. We do not discuss about these models, since they have not open sourced their techniques. 

\subsection*{g1}
g1 \citep{g1} may be the earliest project that attempts to reimplement o1, they take the approach of prompt engineering. They prompt LLM to self-reflect and propose multiple solutions to clone the behavior of o1.

\subsection*{Thinking Claude}
Thinking Claude \citep{thinking-calude} works similarly to g1, which prompts LLM with more complex and fine-grained actions, like problem analysis and progress tracking. Both g1 and Thinking Claude can reshape the behavior of LLM to be like that of o1, but they have not validated their prompts on reasoning benchmarks. 

\subsection*{Open-o1}
Open-o1 \citep{open-o1} provides a SFT dataset, where each response contains long cot. But it is not clear where does these data come from. We suspect that these data are coming from human experts or a strong LLM. Open-o1 found that training LLama-3-8B and Qwen-7b on their dataset can not only shape the style of model response to be like o1, but also improve the model performance on reasoning benchmarks.

\subsection*{o1 Journey}
The o1 journey \citep{o1-journey1,o1-journey2} is outlined in two technical reports. 

\paragraph{Part 1} In Part 1 \citep{o1-journey1}, the tree data generated through beam search is traversed, with specific nodes refined by GPT-4 and then used in supervised fine-tuning. The examples presented in the paper highlight the model's self-reflection capabilities, which comes from the refinement of GPT-4. The approach adopted in Part 1 can be described as expert iteration, where SFT is applied to the data generated via search. Part 1 also compares PRM from o1-mini anotation versus Math-Shepherd~\citep{MathShepherd}, they found that o1-mini outperforms Math-Shepherd.

\paragraph{Part 2} Part 2 \citep{o1-journey2} of the o1 journey introduces a radically different approach. While Part 1 focuses on reinforcement learning, Part 2 attempts to distill o1-mini. Although o1-mini conceals the chain-of-thought (CoT) and only outputs the summary of CoT, Part 2 tries to recover the hidden CoT by prompt o1-mini to augment the summary. Through distillation, they found that Qwen-72B outperforms o1-preview on AIME. However, it does not mean that distillation enables the student model to outperform the teacher model, since o1-mini also surpasses o1-preview on AIME. 

\subsection*{Open-Reasoner}
The framework of Open-Reasoner \citep{open-reasoner} is akin to AlphaGo \citep{AlphaGo}, utilizing reinforcement learning to enhance the model's performance. During the testing phase, Monte Carlo Tree Search (MCTS) is employed to identify optimal solutions. This search algorithm is applied exclusively during testing, while the training data is derived by sampling with the current policy. Additionally, Open-Reasoner employs a method similar to Math-Shepherd \citep{MathShepherd} for training the reward model.

\subsection*{Slow Thinking with LLMs}
Similarly to the o1 journey, Slow Thinking with LLMs is alos outlined in two technical reports \citep{renda-1,renda-2}. 
\paragraph{Part 1}
The part 1 of Slow Thinking with LLMs \citep{renda-1} is similar to OpenReasoner \citep{open-reasoner}, incorporating both reinforcement learning and test-time search. But unlike Open-Reasoner, it employs the DPO instead of PPO algorithm during training. In the testing phase, it also employs the MCTS algorithm for search.

\paragraph{Part 2}
The part 2 of Slow Thinking with LLMs \citep{renda-2} distills from QwQ \citep{qwq} and Deepseek-R1 \citep{shao2024deepseekmathpushinglimitsmathematical} and tries two methods for reinforcement learning: DPO and SFT on data generated from reject sampling. They found that distilling on thousands of examples from QwQ and Deepseek-R1 can significantly bring a substantial improvement on challenging reasoning tasks and reinforcement learning can bring further improvement based on distillation.

\subsection*{Marco-o1}
Marco-o1 \citep{marco-o1} integrates the data from Open-o1 with data generated by the model itself through the MCTS algorithm for SFT training. Marco-o1 demonstrates that prompting the model for self-reflection after each step of the MCTS process enhances the effectiveness of the search.

\subsection*{o1-coder}
o1-coder \citep{o1-coder} attempts to reimplement o1 on the code generation. They train a generator to generate test case for providing outcome reward. With outcome rewards, they use MCTS algorithm to generate code solutions, which are then used for improving policy model through SFT. Their train a PRM following \citet{MathShepherd} which is updated with the improvement of the policy.

\section{Future Directions}
\label{future-directions}

\paragraph{How to adapt o1 to general domains?}
The key to adapt o1 to general domains is the availability of a general reward models. How can we learn a general reward model for general domains? For reasoning tasks, there is typically a standard answer, allowing for the training of an outcome reward model. Using the reward shaping techniques introduced in Section \ref{sec:reward_modeling}, a process reward model can be further trained on top of the outcome reward model. For non-reasoning tasks, such as alignment tasks, obtaining an outcome reward is often challenging. In such cases, methods introduced in Section \ref{sec:reward_modeling}, such as learning from feedback, become crucial. A Bradley-Terry model \citep{ScalableReward} can be trained from preference data, or a reward model can be developed from expert data using inverse reinforcement learning.

\paragraph{How to introduce multiple modality to o1?}
The challenge of incorporating multiple modality into o1 primarily lies in the alignment or grounding between text and other modalities. For example, when the image modality is introduced, the model needs to handle the fine-grained relationships between long CoT and images. When CoT becomes lengthy, establishing such connections becomes more difficult. Therefore, ~\citet{image_cot} incorporated the image modality into the CoT, so that the CoT generated by the model includes both text and images. This approach enhances the fine-grained connections between the generated text and images, but it also introduces new challenges. Introducing information from the image or other modalities into the COT significantly increases its length, leading to increased inference latency. A potential solution to this issue could be to use continious representations to generate CoT, replacing textual and other modality-specific information. 

\paragraph{How to learn and search with a world model?}
OpenAI has outlined a 5-stage roadmap to AGI, with stage 2 focusing on becoming a strong reasoner, stage 3 centered on becoming an agent.The o1 has already reached stage 2, achieving reasoning capabilities on par with human experts. Consequently, o1’s next goal is to progress to stage 3—being able to take actions in real environment and solving real-environment tasks.

The key to extending o1 to real-world environments lies in reward modeling, more specifically in environment or world modeling. This concept of world models was also introduced in Section \ref{sec:reward_modeling}. When a simulator of the real environment is established, the agent can interact with the world model rather than directly with the environment. The world model plays a vital role in both training and testing the agent. During training, interacting with the world model is more efficient than direct interaction with the environment. During testing, the agent can leverage the world model for planning or searching, identifying the optimal strategy before executing actions in the real environment. Without a world model, it is impossible for the model to perform search or planning, as the real environment is not time-reversible. For example, in the game of Go, once a move is made, it cannot be undone.

\section{Conclusion}
In this paper, we present a roadmap for reproducing o1 from the perspective of reinforcement learning, emphasizing key components such as policy initialization, reward design, search, and learning. We offer a comprehensive survey of these components and show that existing open-source projects that attempt to reproduce o1 are variations of our roadmap. Finally, we hope this roadmap inspire further research on overcoming the challenges involved in reproducing o1.
\bibliography{colm2024_conference}
\bibliographystyle{colm2024_conference}

\appendix
\end{document}